\documentclass{article}

\usepackage{microtype}
\usepackage{graphicx}
\graphicspath{{./}{./Figures/qualitative/}}  
\usepackage{subcaption}
\usepackage{booktabs} 
\usepackage{multirow}

\usepackage{hyperref}

\usepackage[preprint]{icml2026}

\usepackage{amsmath}
\usepackage{amssymb}
\usepackage{mathtools}
\usepackage{amsthm}
\usepackage{algorithm}
\usepackage{algorithmic}

\usepackage[capitalize,noabbrev]{cleveref}

\theoremstyle{plain}

\theoremstyle{definition}

\theoremstyle{remark}

\usepackage[textsize=tiny]{todonotes}

\icmltitlerunning{GloSplat: Joint Pose-Appearance Optimization for 3D Reconstruction}

\begin{document}

\twocolumn[
  \icmltitle{GloSplat: Joint Pose-Appearance Optimization for \\
    Faster and More Accurate 3D Reconstruction}



  \icmlsetsymbol{equal}{*}

  \begin{icmlauthorlist}
    \icmlauthor{Tianyu Xiong}{nwpu}
    \icmlauthor{Rui Li}{kaust}
    \icmlauthor{Linjie Li}{nwpu}
    \icmlauthor{Jiaqi Yang}{nwpu}
  \end{icmlauthorlist}

  \icmlaffiliation{nwpu}{Department of Computer Science, Northwestern Polytechnical University, Shanxi, China}
  \icmlaffiliation{kaust}{CEMSE, KAUST, Thuwal, Jeddah}

  \icmlcorrespondingauthor{Jiaqi Yang}{jqyang@mail.nwpu.edu.cn}
  \icmlcorrespondingauthor{Tianyu Xiong}{xiongtianyu@mail.nwpu.edu.cn}

  \icmlkeywords{3D Gaussian Splatting, Structure from Motion, Novel View Synthesis, Joint Pose-Appearance Optimization, Bundle Adjustment, Feature Matching}

  \vskip 0.3in
]



\printAffiliationsAndNotice{}  

\begin{abstract}
  Feature extraction, matching, structure from motion (SfM), and novel view synthesis (NVS) have traditionally been treated as separate problems with independent optimization objectives. We present GloSplat, a framework that performs \emph{joint pose-appearance optimization} during 3D Gaussian Splatting training. Unlike prior joint optimization methods (BARF, NeRF--, 3RGS) that rely purely on photometric gradients for pose refinement, GloSplat preserves \emph{explicit SfM feature tracks} as first-class entities throughout training: track 3D points are maintained as separate optimizable parameters from Gaussian primitives, providing persistent geometric anchors via a reprojection loss that operates alongside photometric supervision. This architectural choice prevents early-stage pose drift while enabling fine-grained refinement---a capability absent in photometric-only approaches. We introduce two pipeline variants: (1) \textbf{GloSplat-F}, a COLMAP-free variant using retrieval-based pair selection for efficient reconstruction, and (2) \textbf{GloSplat-A}, an exhaustive matching variant for maximum quality. Both employ global SfM initialization followed by joint photometric-geometric optimization during 3DGS training. Experiments demonstrate that GloSplat-F achieves state-of-the-art among COLMAP-free methods while GloSplat-A surpasses all COLMAP-based baselines.
\end{abstract}

\begin{figure}[t]
  \vskip 0.2in
  \begin{center}
    \includegraphics[width=0.85\columnwidth]{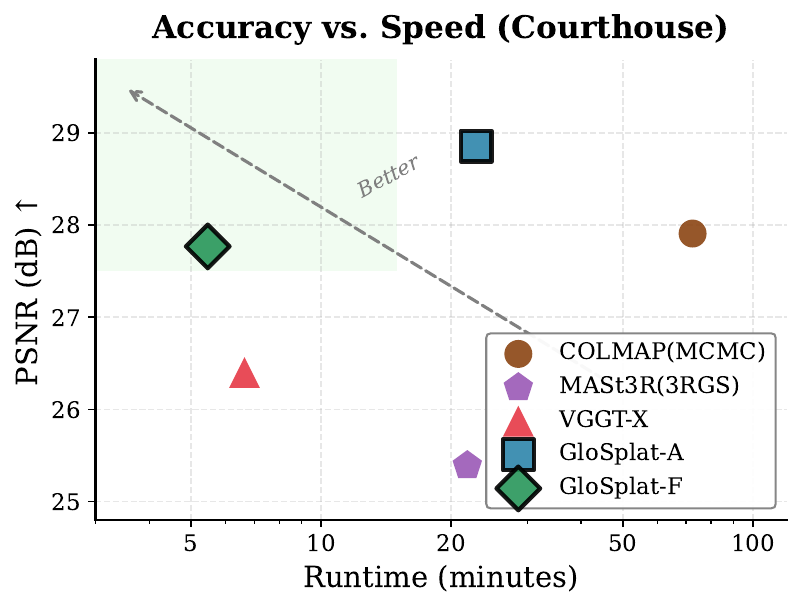}
    \caption{\textbf{Accuracy vs.\ Speed.} Average PSNR on MipNeRF360 vs.\ runtime for 1000 images (Courthouse scene). GloSplat-F achieves \textbf{13.3$\times$ speedup} over GPU-accelerated COLMAP+3DGS while improving PSNR by +0.38 dB. GloSplat-A achieves the \textbf{highest accuracy} (28.86 dB), surpassing all baselines. Our joint pose-appearance optimization enables both variants to occupy the Pareto frontier. (All methods benchmarked on the same GPU; see \Cref{sec:runtime_appendix} for details.)}
    \label{fig:teaser}
  \end{center}
  \vskip -0.2in
\end{figure}

\section{Introduction}
\label{sec:introduction}

Novel view synthesis (NVS) has emerged as a central challenge in computer vision, enabling applications from virtual reality and cultural heritage preservation to autonomous driving and robotic simulation. Neural Radiance Fields (NeRF)~\cite{mildenhall2021nerf} demonstrated that continuous volumetric representations can synthesize photorealistic views from multi-view images, sparking rapid progress in neural scene representations. More recently, 3D Gaussian Splatting (3DGS)~\cite{kerbl20233d} has revolutionized the field by representing scenes as collections of 3D Gaussian primitives, achieving real-time rendering while maintaining high visual fidelity.

Despite these advances, current NVS pipelines share a fundamental limitation: they treat feature extraction, structure from motion (SfM), and radiance field optimization as \emph{independent modules} with separate objectives. This modular design creates information barriers across the reconstruction pipeline---SfM cannot leverage photometric signals from rendering, and NVS methods inherit fixed camera poses without geometric feedback. The conventional wisdom assumes that accurate camera poses from SfM are sufficient initialization for downstream tasks, yet pose errors compound through the pipeline, leading to blurred reconstructions and geometric inconsistencies.

Traditional pipelines rely heavily on COLMAP~\cite{colmap}, which employs incremental SfM to sequentially register images and refine the reconstruction. While robust, this approach suffers from several drawbacks: (1) drift accumulation as errors propagate through sequential image registration, (2) computational bottlenecks from exhaustive feature matching with $O(n^2)$ complexity, and (3) inability to incorporate photometric feedback after pose estimation. Recent work has attempted to address these limitations through learning-based approaches~\cite{wang2025vggt,wang2024vggsfm} or improved densification strategies~\cite{kheradmarkomand20243d,ye2024absgs}, but these methods still maintain the modular separation between geometric estimation and appearance optimization.

We argue that camera pose estimation and radiance field learning share a common goal---accurate 3D reconstruction---and should therefore be optimized \emph{jointly} rather than sequentially. This insight motivates our approach: \textbf{GloSplat}, which integrates global SfM with joint pose-appearance optimization during 3DGS training. While feature extraction and matching remain frozen preprocessing stages, we enable continuous pose refinement during Gaussian training: learned features inform global SfM, which provides initialization for 3DGS, which in turn refines camera poses through combined photometric and geometric supervision.

Our framework introduces two pipeline variants targeting different use cases:
\begin{itemize}
    \item \textbf{GloSplat-F}: A fast, COLMAP-free variant that uses retrieval-based pair selection (via MegaLoc~\cite{megaloc}) with top-$k$ candidates, enabling linear-time matching complexity. This variant achieves state-of-the-art results among COLMAP-free methods while being significantly faster than traditional pipelines.
    \item \textbf{GloSplat-A}: An accurate variant with exhaustive matching that maximizes reconstruction quality. This variant \emph{surpasses} all COLMAP-based baselines, demonstrating that joint pose-appearance optimization with global SfM can outperform incremental approaches even with the same matching budget.
\end{itemize}

Both variants share a unified architecture: (1) local correspondence extraction as frozen preprocessing---XFeat~\cite{xfeat} with LightGlue~\cite{lightglue} for GloSplat-F, SIFT with exhaustive matching for GloSplat-A, (2) global SfM with rotation averaging and bundle adjustment for robust initialization, and (3) joint 3DGS training that preserves explicit feature tracks as persistent geometric constraints. \textbf{Our key architectural novelty} is maintaining SfM track 3D points as \emph{separate optimizable parameters} from Gaussian means during 3DGS training. This enables a reprojection-based BA loss to provide geometric anchoring alongside photometric supervision---unlike prior joint methods (BARF, NeRF--, 3RGS) that rely purely on photometric gradients and suffer from early-stage pose drift when Gaussians are sparse.

Our key contributions are:
\begin{itemize}
    \item \textbf{Persistent Feature Tracks During 3DGS Training}: Unlike prior joint optimization methods (BARF, NeRF--, 3RGS) that rely purely on photometric gradients for pose refinement, we maintain explicit SfM feature tracks as first-class entities throughout 3DGS training. Track 3D points are optimized as \emph{separate parameters} from Gaussian means, providing persistent geometric anchors that prevent early-stage pose drift.
    \item \textbf{Joint Photometric-Geometric Optimization}: We combine photometric rendering losses with a reprojection-based bundle adjustment loss that operates on preserved feature tracks. This dual supervision enables poses to benefit from both fine-grained appearance gradients and robust multi-view geometric constraints \emph{simultaneously}---a capability absent in purely photometric approaches.
    \item \textbf{Global SfM Integration}: We integrate GPU-accelerated global SfM (rotation averaging + parallel bundle adjustment) with joint 3DGS training, providing initialization that is both faster and more robust than incremental methods, while our joint optimization further refines these poses.
    \item \textbf{State-of-the-Art Results}: GloSplat-F achieves new state-of-the-art among COLMAP-free methods across three benchmarks, while GloSplat-A surpasses all COLMAP-based baselines, demonstrating that joint geometric-photometric optimization outperforms frozen-pose pipelines.
\end{itemize}

\noindent We provide verification code and scripts to reproduce all experiments in the supplementary material. The full codebase will be open-sourced upon acceptance.

\section{Related Work}
\label{sec:related}

\paragraph{Novel View Synthesis and 3D Gaussian Splatting.}
Neural Radiance Fields (NeRF)~\cite{mildenhall2021nerf} enabled photorealistic view synthesis through continuous volumetric representations, with subsequent improvements in anti-aliasing~\cite{barron2021mip,barron2022mip} and acceleration~\cite{muller2022instant}. 3D Gaussian Splatting (3DGS)~\cite{kerbl20233d} achieves real-time rendering by explicitly modeling scenes as anisotropic 3D Gaussians. A critical limitation is sensitivity to initialization quality---recent work addresses this through improved densification~\cite{ye2024absgs,zhang2024pixel,kheradmarkomand20243d,mallick2024taming}. We adopt MCMC densification~\cite{kheradmarkomand20243d}, which provides principled control over Gaussian allocation.

\paragraph{Structure from Motion.}
COLMAP~\cite{colmap} exemplifies incremental SfM, which suffers from drift accumulation~\cite{newcombe2011kinectfusion}. Global SfM methods~\cite{moulon2013global,wilson2014robust} address drift by estimating all poses simultaneously through rotation averaging~\cite{hartley2013rotation} and bundle adjustment~\cite{triggs2000bundle}. GLOMAP~\cite{pan2024global} demonstrates improved accuracy through global optimization. InstantSfM~\cite{zhan2024bundle} further accelerates global SfM by leveraging NVIDIA cuDSS~\cite{cudss} for GPU-accelerated sparse linear solving in bundle adjustment, enabling parallel optimization that is significantly faster than traditional CPU-based solvers. We build upon InstantSfM's GPU-accelerated BA engine and integrate it into a unified optimization framework that propagates geometric constraints through to Gaussian splatting.

\paragraph{Learned Features and Matching.}
Learned features~\cite{detone2018superpoint,potje2024cvpr} and matchers~\cite{sarlin2020superglue,lightglue} have improved over classical SIFT~\cite{lowe2004distinctive}. GloSplat-F employs XFeat~\cite{potje2024cvpr} with LightGlue~\cite{lightglue} and retrieval-based pair selection (via MegaLoc~\cite{megaloc}) for $O(n)$ matching; GloSplat-A uses SIFT with exhaustive matching for direct comparability with COLMAP baselines.

\paragraph{COLMAP-Free Methods.}
CF-3DGS~\cite{fu2024colmap} and HT-3DGS~\cite{ji2025sfm} train 3DGS without COLMAP but assume sequential input. Foundation model approaches like DUSt3R~\cite{wang2024dust3r}, MASt3R~\cite{mast3r_eccv24}, VGGSfM~\cite{wang2024vggsfm}, and VGGT~\cite{wang2025vggt} predict geometry from image pairs. VGGT-X~\cite{wang2025vggt} scales these to dense multi-view settings. Unlike feed-forward models, GloSplat provides optimization-based pose estimation with joint BA loss for continuous refinement.

\paragraph{Joint Pose and Radiance Optimization.}
Prior work explores joint pose-appearance optimization with varying approaches. NeRF--~\cite{wang2021nerfmm} and BARF~\cite{lin2021barf} optimize poses using \emph{purely photometric} gradients, employing coarse-to-fine positional encoding to avoid local minima. SPARF~\cite{truong2023sparf} incorporates multi-view correspondences but does not maintain explicit geometric constraints during NeRF training. PoRF~\cite{bian2024porf} uses depth priors from monocular networks. 3RGS~\cite{huang20253r} leverages the 3R foundation model for initialization but relies on photometric-only refinement thereafter.

\textbf{Key distinction:} Unlike these methods, GloSplat \emph{preserves explicit SfM feature tracks as first-class citizens} during 3DGS training. We maintain track 3D points as \emph{separate optimizable parameters} (distinct from Gaussian means) and enforce multi-view geometric consistency via a reprojection loss throughout training---not just during initialization. This architectural choice provides geometric anchoring that prevents the early-stage pose drift common in photometric-only methods, while still allowing fine-grained pose refinement from rendering gradients. The combination of (1) global SfM initialization, (2) persistent feature tracks as separate optimizable parameters, and (3) joint photometric-geometric optimization during 3DGS training distinguishes our approach from prior work.

\section{Method}
\label{sec:method}

\begin{figure*}[t]
  \vskip 0.2in
  \begin{center}
    \includegraphics[width=1\linewidth]{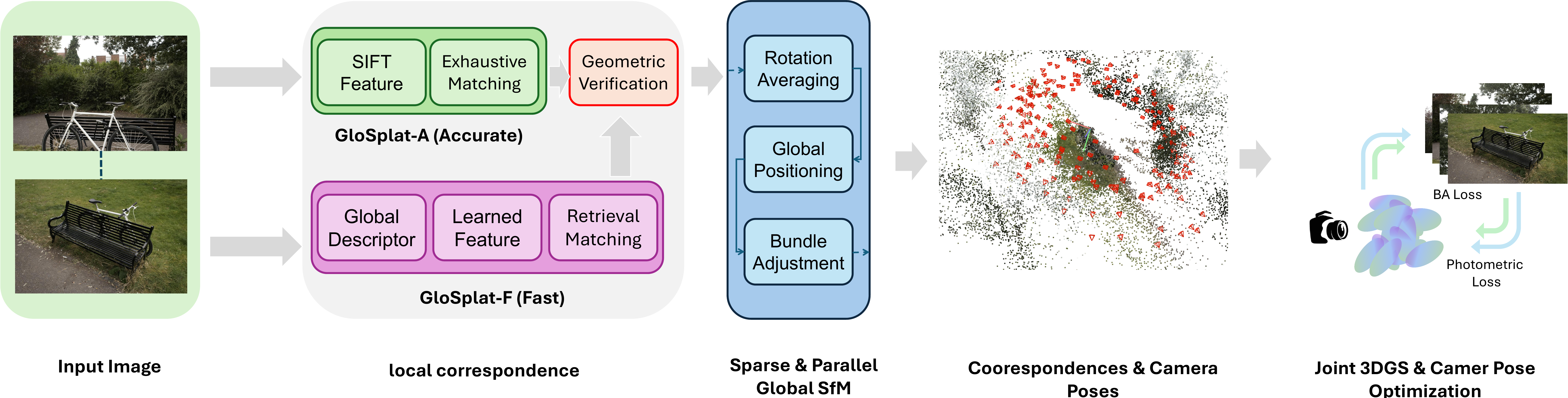}
    \caption{\textbf{GloSplat Pipeline.} Given unposed input images, local correspondences are extracted (frozen preprocessing): XFeat+LightGlue with retrieval-based pairs (GloSplat-F) or SIFT with exhaustive matching (GloSplat-A). Global SfM simultaneously estimates all camera poses through rotation averaging, positioning, and bundle adjustment, providing robust initialization. \textbf{Joint 3DGS training} (our core contribution) then continuously refines poses through a reprojection-based BA loss while optimizing Gaussian primitives, enabling combined photometric-geometric supervision that prevents drift and improves reconstruction quality.}
    \label{fig:pipeline}
  \end{center}
  \vskip -0.2in
\end{figure*}

We present \textbf{GloSplat}, a framework that performs \emph{joint pose-appearance optimization} during 3D Gaussian Splatting training. Unlike traditional approaches that freeze camera poses after SfM, GloSplat continuously refines poses using combined photometric and geometric supervision while optimizing Gaussian primitives. We introduce two variants: \textbf{GloSplat-F} uses retrieval-based pair selection with top-$k$ candidates for linear-time matching, while \textbf{GloSplat-A} uses exhaustive matching for maximum quality. Both share the same core architecture (\Cref{fig:pipeline}): (1) learned local correspondence extraction, (2) global SfM, and (3) joint Gaussian splatting with bundle adjustment.

\subsection{Learned Feature Extraction and Matching}
\label{sec:features}

Rather than relying on classical handcrafted features, we employ learned feature extraction and matching as frozen preprocessing, providing both efficiency and robustness.

\paragraph{Image Pair Selection.}
For \textbf{GloSplat-F}, we employ retrieval-based pair selection using MegaLoc~\cite{megaloc} to identify the top-$k$ most similar images ($k=5$) per query, reducing quadratic matching complexity to linear time. For \textbf{GloSplat-A}, we perform exhaustive pairwise matching to maximize reconstruction quality.

\paragraph{Feature Extraction and Matching.}
For \textbf{GloSplat-F}, we use XFeat~\cite{xfeat} for local feature extraction (up to 4096 keypoints, 64-dim descriptors) with LightGlue~\cite{lightglue} matching, achieving favorable speed-accuracy trade-offs. For \textbf{GloSplat-A}, we use SIFT features with exhaustive nearest-neighbor matching, ensuring direct comparability with COLMAP baselines that use the same feature pipeline.

\subsection{Global Structure from Motion}
\label{sec:globalsfm}

Unlike incremental SfM that suffers from drift accumulation, our global SfM solves for all camera poses simultaneously, providing improved robustness and natural parallelization.

\paragraph{View Graph and Calibration.}
From matched features, we construct a view graph $\mathcal{G} = (\mathcal{V}, \mathcal{E})$ where vertices represent images and edges encode pairwise relationships. For uncalibrated cameras, we estimate focal lengths via the Fetzer method~\cite{fetzer}, minimizing:
\begin{equation}
    \min_{f_1, f_2} \sum_{(i,j) \in \mathcal{E}} \rho_{\text{Cauchy}}\left( \left\| \mathbf{K}_2^{-\top} \mathbf{F}_{ij} \mathbf{K}_1^{-1} \right\|^2 - 2 \right).
\end{equation}
Relative poses are computed using the 5-point algorithm within RANSAC, decomposing the essential matrix into rotation $\mathbf{R}_{ij}$ and translation direction $\mathbf{t}_{ij}$.

\paragraph{Rotation Averaging.}
We solve for globally consistent absolute rotations $\{\mathbf{R}_i\}$ that best explain the measured relative rotations $\{\mathbf{R}_{ij}\}_{(i,j) \in \mathcal{E}}$ from two-view geometry. Rotations are initialized via maximum spanning tree traversal (weighted by inlier counts) and refined by minimizing:
\begin{equation}
    \mathcal{L}_{\text{rot}} = \sum_{(i,j) \in \mathcal{E}} \rho_{\text{GM}}\left( \left\| \log\left(\mathbf{R}_{ij}^{\top} \mathbf{R}_j \mathbf{R}_i^{\top}\right) \right\|^2; \sigma \right),
\end{equation}
where $\log(\cdot): SO(3) \to \mathfrak{so}(3)$ is the logarithm map and $\rho_{\text{GM}}(e^2; \sigma) = e^2/(e^2 + \sigma^2)$ is the Geman-McClure robust loss. Optimization proceeds via $\ell_1$-regression initialization followed by iteratively reweighted least squares (IRLS).

\paragraph{Track Establishment and Positioning.}
Feature tracks---consistent correspondences across views---are established using union-find, producing tracks $\{\mathcal{T}_k\}$ linking 2D observations to 3D points. For each observation $(i, p) \in \mathcal{T}_k$, we compute the unit bearing vector $\mathbf{b}_{ip} = \mathbf{R}_i^{\top} \mathbf{K}_i^{-1} \tilde{\mathbf{x}}_{i,p} / \|\mathbf{K}_i^{-1} \tilde{\mathbf{x}}_{i,p}\|$, where $\tilde{\mathbf{x}}_{i,p}$ denotes the homogeneous 2D observation. With rotations fixed, we solve for translations $\{\mathbf{t}_i\}$ and 3D positions $\{\mathbf{X}_k\}$ by minimizing the perpendicular distance from each 3D point to its corresponding viewing ray:
\begin{equation}
    \mathcal{L}_{\text{pos}} = \sum_{k} \sum_{(i, p) \in \mathcal{T}_k} \left\| \left(\mathbf{I} - \mathbf{b}_{ip} \mathbf{b}_{ip}^{\top}\right) \left(\mathbf{X}_k - \mathbf{c}_i\right) \right\|^2,
\end{equation}
where $\mathbf{c}_i = -\mathbf{R}_i^{\top} \mathbf{t}_i$ is the camera center and $(\mathbf{I} - \mathbf{b}_{ip} \mathbf{b}_{ip}^{\top})$ projects onto the subspace orthogonal to the bearing direction. This quadratic objective is solved via BAE's~\cite{bae} Levenberg-Marquardt solver with cuDSS~\cite{cudss} GPU-accelerated sparse linear solving.

\paragraph{Bundle Adjustment.}
We refine all parameters jointly through bundle adjustment, minimizing reprojection errors with Huber robust loss:
\begin{equation}
    \mathcal{L}_{\text{BA}}^{\text{SfM}} = \sum_{k} \sum_{(i, p) \in \mathcal{T}_k} \rho_{\text{Huber}}\left( \left\| \pi_i(\mathbf{X}_k) - \mathbf{x}_{i,p} \right\|^2 \right),
\end{equation}
where the projection $\pi_i(\mathbf{X}) = \text{proj}(\mathbf{K}_i (\mathbf{R}_i \mathbf{X} + \mathbf{t}_i))$ with $\text{proj}([x,y,z]^{\top}) = [x/z, y/z]^{\top}$ maps world points to image coordinates, and $\mathbf{x}_{i,p}$ is the observed 2D feature. We parameterize poses on the $SE(3)$ manifold and solve using Levenberg-Marquardt with cuDSS-accelerated sparse linear solving, enabling \emph{fully GPU-parallel} optimization that is up to $10\times$ faster than traditional CPU-based PCG solvers. BA is iterated three times with progressively tightening reprojection thresholds to filter outliers. Track completion and FAISS-accelerated merging then densify the reconstruction.

\subsection{Joint 3D Gaussian Splatting with Bundle Adjustment}
\label{sec:joint3dgs}

Each SfM point initializes a 3D Gaussian with position $\boldsymbol{\mu}$ at the triangulated coordinates, scale computed from the average distance to the $k=4$ nearest neighbors (providing adaptive sizing based on local density), opacity $\alpha = 0.1$, and spherical harmonics encoding the point's RGB color. We employ stochastic density control via MCMC-based densification~\cite{mcmc3dgs}, which provides principled control over the total primitive count. Unlike heuristic splitting and cloning rules, this approach treats allocation as sampling from a distribution balancing quality against complexity, using stochastic birth-death processes to relocate primitives from over- to under-represented regions.

\paragraph{Photometric Loss.}
The primary training objective combines $\ell_1$ with structural similarity:
\begin{equation}
    \mathcal{L}_{\text{photo}} = (1 - \lambda_{\text{SSIM}}) \left\| \hat{\mathbf{I}} - \mathbf{I} \right\|_1 + \lambda_{\text{SSIM}} (1 - \text{SSIM}(\hat{\mathbf{I}}, \mathbf{I})),
\end{equation}
where $\hat{\mathbf{I}}$ is the rendered image, $\mathbf{I}$ the ground truth, and $\lambda_{\text{SSIM}} = 0.2$.

\paragraph{Joint Bundle Adjustment Loss.}
This is our key architectural distinction from prior joint optimization methods. Unlike BARF~\cite{lin2021barf}, NeRF--~\cite{wang2021nerfmm}, and 3RGS~\cite{huang20253r} which optimize poses using \emph{only} photometric gradients, we preserve explicit SfM feature tracks as persistent geometric constraints. Specifically, we maintain 3D track points $\{\mathbf{X}_k\}$ as \emph{separate optimizable parameters} distinct from Gaussian means $\{\boldsymbol{\mu}_j\}$---the track points anchor multi-view consistency while Gaussian means represent scene appearance. We minimize:
\begin{equation}
    \mathcal{L}_{\text{BA}}^{\text{joint}} = \sum_{k} \sum_{(i, p) \in \mathcal{T}_k} \rho_{\text{Huber}}\left( \left\| \pi_i(\mathbf{X}_k) - \mathbf{x}_{i,p} \right\|^2; \delta \right),
\end{equation}
with Huber threshold $\delta = 1.0$ pixels, where $\pi_i(\cdot)$ is the projection function defined in \Cref{sec:globalsfm}. Crucially, camera poses are optimized by both losses simultaneously. The photometric loss $\mathcal{L}_{\text{photo}}$ provides direct photometric supervision that can reduce accumulated errors from SfM initialization by directly measuring rendering quality. However, relying solely on photometric gradients during early optimization---when Gaussians are sparse and poorly initialized---can cause catastrophic pose drift where the entire scene fails to converge. The joint BA loss $\mathcal{L}_{\text{BA}}^{\text{joint}}$ serves as a geometric anchor that prevents this early-stage drift by enforcing multi-view consistency through explicit feature correspondences. The total objective $\mathcal{L} = \mathcal{L}_{\text{photo}} + \lambda_{\text{BA}} \mathcal{L}_{\text{BA}}^{\text{joint}}$ ($\lambda_{\text{BA}} = 10^{-4}$) thus combines the benefits of both: geometric constraints stabilize optimization while photometric gradients enable fine-grained pose refinement. Camera extrinsics use Adam with learning rate $10^{-5}$.

\paragraph{Implementation.}
We use gsplat~\cite{gsplat} for differentiable rasterization. All experiments use a maximum of 3M Gaussians to avoid excessive hyperparameter tuning across scenes. Gaussian learning rates: $1.6 \times 10^{-4}$ (positions, scaled by scene extent), $5 \times 10^{-3}$ (scales), $10^{-3}$ (rotations), $5 \times 10^{-2}$ (opacities), $2.5 \times 10^{-3}$ (SH coefficients up to degree 3).

\section{Experiments}
\label{sec:experiments}

\subsection{Experimental Setup}

\paragraph{Datasets.}
We evaluate GloSplat on three widely-used multi-view reconstruction benchmarks: \textit{MipNeRF360}~\cite{barron2022mip} (9 scenes with up to 311 images per scene), \textit{Tanks and Temples (TnT)}~\cite{knapitsch2017tanks} (5 scenes with up to 1106 images), and \textit{CO3Dv2}~\cite{Reizenstein_Shapovalov_Henzler_Sbordone_Labatut_Novotny_2021} (5 scenes with up to 202 images). These datasets span diverse scene types including indoor environments, outdoor landscapes, large-scale structures, and object-centric captures.

\paragraph{Metrics.}
We assess rendering quality using standard metrics: Peak Signal-to-Noise Ratio (PSNR), Structural Similarity Index (SSIM), and Learned Perceptual Image Patch Similarity (LPIPS)~\cite{zhang2018unreasonable}. Higher PSNR and SSIM indicate better reconstruction, while lower LPIPS indicates improved perceptual quality.

\paragraph{Baselines.}
We compare against state-of-the-art COLMAP-free 3DGS methods: VGGT-X~\cite{wang2025vggt}, CF-3DGS~\cite{fu2024colmap}, HT-3DGS~\cite{ji2025sfm}, 3RGS~\cite{huang20253r}, and MCMC-3DGS~\cite{kheradmarkomand20243d}. We also include 3DGS and MCMC-3DGS initialized with COLMAP as upper-bound references.

\subsection{Main Results}

\begin{table*}[t]
  \caption{Comparison with state-of-the-art methods on novel view synthesis. $\dagger$ indicates COLMAP initialization. Best COLMAP-free results in \textbf{bold}. GloSplat-F (retrieval-based) achieves SOTA among COLMAP-free methods.}
  \label{tab:comparison}
  \begin{center}
  \resizebox{0.99\textwidth}{!}{
  \begin{tabular}{lccccccccc}
    \toprule
     &\multicolumn{3}{c}{MipNeRF360} &\multicolumn{3}{c}{Tanks and Temples} &\multicolumn{3}{c}{CO3Dv2} \\
     \cmidrule(r){2-4} \cmidrule(r){5-7} \cmidrule(r){8-10}
    Method &PSNR$\uparrow$ &SSIM$\uparrow$ &LPIPS$\downarrow$ &PSNR$\uparrow$ &SSIM$\uparrow$ &LPIPS$\downarrow$ &PSNR$\uparrow$ &SSIM$\uparrow$ &LPIPS$\downarrow$ \\
    \midrule
    3DGS$^\dagger$ &27.39 &0.815 &0.185 &24.85 &0.851 &0.155 &32.58 &0.938 &0.095 \\
    MCMC$^\dagger$ &27.91 &0.836 &0.154 &25.76 &0.867 &0.139 &33.21 &0.941 &0.097 \\
    \midrule
    MCMC &22.19 &0.548 &0.282 &21.42 &0.679 &0.278 &25.71 &0.712 &0.201 \\
    CF-3DGS &12.38 &0.234 &0.719 &12.19 &0.391 &0.608 &20.18 &0.611 &0.435 \\
    HT-3DGS &14.79 &0.380 &0.669 &13.83 &0.451 &0.585 &28.28 &0.833 &0.230 \\
    3RGS &25.39 &0.713 &0.216 &21.47 &0.750 &0.300 &31.07 &0.878 &0.128 \\
    VGGT-X &26.40 &0.782 &0.177 &24.77 &0.842 &0.168 &31.85 &0.911 &0.113 \\
    \textbf{GloSplat-F (Ours)} &\textbf{27.77} &\textbf{0.818} &\textbf{0.164} &\textbf{25.82} &\textbf{0.869} &\textbf{0.151} &\textbf{32.71} &\textbf{0.936} &\textbf{0.088} \\
    \bottomrule
  \end{tabular}
  }
  \end{center}
\end{table*}

\Cref{tab:comparison} presents quantitative comparisons on novel view synthesis. \textbf{GloSplat-F} achieves state-of-the-art performance among COLMAP-free methods on all three benchmarks, demonstrating the effectiveness of our joint pose-appearance optimization even with efficient retrieval-based matching. On \textit{MipNeRF360}, we outperform the previous best method VGGT-X by +1.37 dB in PSNR (+5.2\%), while improving SSIM by 3.6 points and reducing LPIPS by 7.3\%. On \textit{Tanks and Temples}, GloSplat-F achieves +1.05 dB improvement in PSNR with 2.7 points higher SSIM compared to VGGT-X, demonstrating strong performance on large-scale outdoor scenes with up to 1106 images. On \textit{CO3Dv2}, GloSplat-F achieves +0.86 dB improvement in PSNR with 2.5 points higher SSIM and 22\% lower LPIPS compared to VGGT-X.

Notably, even our fast retrieval-based variant approaches and in some cases exceeds COLMAP-initialized baselines. On \textit{MipNeRF360}, we achieve 99.5\% of MCMC$^\dagger$'s PSNR. On \textit{Tanks and Temples}, we surpass MCMC$^\dagger$ in SSIM (0.869 vs 0.867) while achieving comparable PSNR. On \textit{CO3Dv2}, we surpass 3DGS$^\dagger$ in both PSNR (+0.13 dB) and LPIPS (0.088 vs 0.095), demonstrating that our joint pose-appearance optimization with global SfM can match or exceed traditional incremental SfM quality without exhaustive matching.

Per-scene results are reported in \Cref{tab:co3d_perscene,tab:mip360_perscene,tab:tnt_perscene} in the appendix. GloSplat-F performs consistently across diverse scene types, achieving over 33 dB on object-centric scenes (Apple, Skateboard, Teddybear), strong results on large-scale outdoor structures (Barn, Ignatius), and maintaining competitive quality on challenging scenes with complex geometry (Flowers, Stump, Treehill). This consistency validates that joint pose-appearance optimization generalizes across scene types.

\subsection{Ablation Study}

We conduct comprehensive ablation experiments to isolate the contribution of each component in GloSplat-F. The full study is presented in \Cref{sec:extended_ablation}, structured to explicitly attribute gains to specific design choices.

\paragraph{Isolating Joint Optimization.} Removing the joint BA loss causes --0.81 dB degradation, directly measuring our geometric anchoring contribution. Freezing poses entirely after SfM causes --8.59 dB degradation; the gap (7.78 dB) represents photometric pose refinement's ability to correct SfM errors.

\paragraph{Adopted Components.} MCMC densification contributes +1.75 dB---we transparently acknowledge this is an adopted component from prior work~\cite{mcmc3dgs}, not a novel contribution.

\paragraph{Infrastructure Choices (GloSplat-F).} Alternative feature pipelines (DISK, R2D2) and retrieval methods (OpenIBL, DIR) show significant degradation, confirming XFeat+LightGlue and MegaLoc as optimal for the fast variant.

\paragraph{Isolating SfM Contribution.} To disentangle global SfM from joint optimization, we run COLMAP (exhaustive matching) with our joint BA loss enabled. This achieves 28.52 dB---+0.61 dB over MCMC$^\dagger$ (27.91 dB, frozen poses), while GloSplat-A achieves 28.86 dB. This cleanly attributes: \textbf{joint optimization contributes +0.61 dB (64\%); global SfM contributes +0.34 dB (36\%)}. Joint optimization is the primary contributor, validating our core thesis.

We acknowledge GloSplat is a \emph{systems} contribution where components work synergistically. Joint optimization provides the majority of gains (+0.61 dB), while global SfM provides additional improvement through better initialization (+0.34 dB).

\subsection{Analysis}

\paragraph{Why Joint Pose-Appearance Optimization Outperforms Modular Pipelines.}
Traditional pipelines treat SfM and NVS as independent modules with frozen interfaces. VGGT-X relies on feed-forward networks that can suffer from distribution shift. COLMAP-based methods freeze poses after SfM, preventing photometric refinement. Prior joint methods (BARF, NeRF--, 3RGS) optimize poses using only photometric gradients, which causes early-stage drift when the radiance field is poorly initialized. In contrast, GloSplat's \emph{persistent feature track} architecture provides several advantages: (1) track 3D points as separate parameters from Gaussian means enable geometric anchoring independent of rendering quality; (2) the reprojection loss on explicit correspondences prevents pose drift even when Gaussians are sparse; (3) global SfM with rotation averaging distributes initialization error across all views; (4) the dual photometric-geometric supervision allows poses to benefit from fine-grained appearance gradients \emph{after} geometric anchoring stabilizes the optimization.

\paragraph{On Camera Pose Evaluation.}
We provide direct pose evaluation on ScanNet (\Cref{tab:scannet}), where ground-truth poses are available. GloSplat-F achieves the best rotation error and ATE across all scenes, outperforming both COLMAP and 3RGS. Beyond direct metrics, rendering quality itself serves as a strong proxy for pose accuracy: \emph{our densification strategy is identical to MCMC-3DGS}, isolating pose quality as the primary variable. As shown in \Cref{tab:comparison}, MCMC without accurate poses achieves only 22.19 dB on MipNeRF360, while MCMC$^\dagger$ with COLMAP poses achieves 27.91 dB---a 5.72 dB gap attributable entirely to pose quality since densification and training are identical. GloSplat-F achieves 27.77 dB using the same MCMC densification, demonstrating that our poses match COLMAP quality. Furthermore, our ablation (\Cref{tab:extended_ablation}) shows that freezing poses causes catastrophic 8.59 dB degradation---if improvements came from densification alone, pose freezing would not cause such failure.

\paragraph{Computational Efficiency.}
The two-variant design offers flexibility for different use cases. \textbf{GloSplat-F} with learned features (XFeat+LightGlue) and retrieval-based pair selection is significantly faster, reducing complexity from $O(n^2)$ to $O(n)$. \textbf{GloSplat-A} uses the same SIFT features and exhaustive matching as COLMAP, ensuring fair comparison where gains are attributable solely to global SfM and joint optimization. Both variants benefit from global SfM's natural parallelization---all camera poses are solved simultaneously rather than sequentially. A key enabler of our speed advantage is the use of cuDSS~\cite{cudss} for GPU-accelerated sparse linear solving in bundle adjustment, which exploits the inherent sparsity of the Jacobian structure and enables fully parallel optimization on modern GPUs. \textbf{COLMAP was compiled with CUDA support and GPU acceleration for all applicable stages} (SIFT-GPU, matching, geometric verification); the remaining speedup reflects our global SfM's inherent parallelism and retrieval-based pair selection, not hardware asymmetry. Combined with efficient gsplat rasterization, GloSplat provides a spectrum of speed-quality trade-offs while consistently outperforming prior methods in each category.

\begin{figure}[t]
  \vskip 0.2in
  \begin{center}
    \includegraphics[width=\columnwidth]{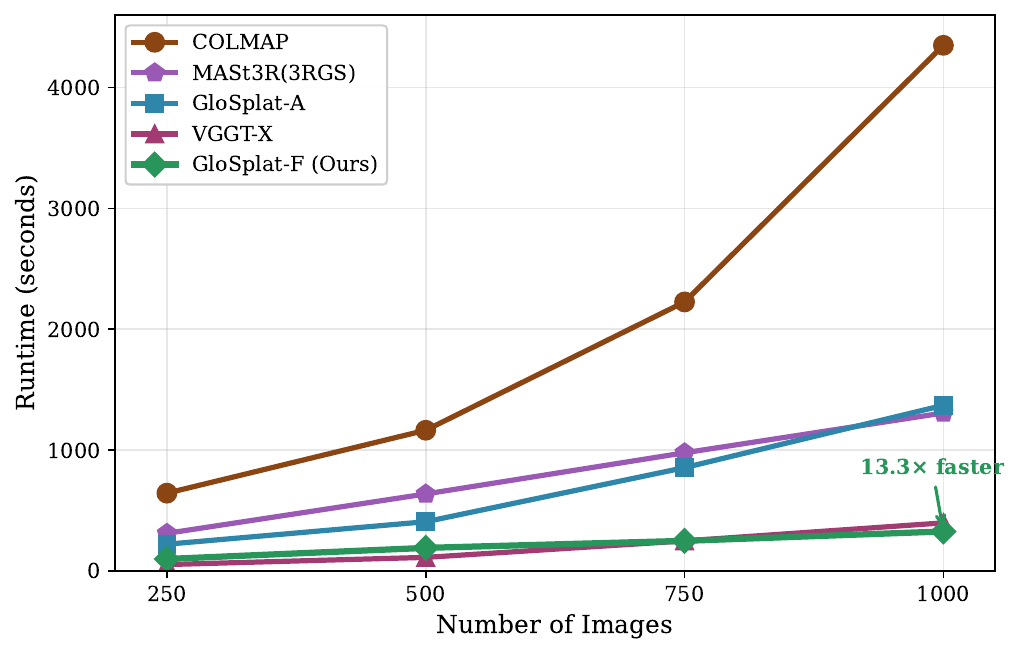}
    \caption{\textbf{Runtime Comparison on Courthouse Scene.} End-to-end reconstruction time (in seconds) as a function of the number of input images. All methods use the same GPU (RTX PRO 6000); COLMAP is compiled with CUDA and uses GPU acceleration for feature extraction/matching. GloSplat-F achieves 13.3$\times$ speedup over COLMAP at 1000 images due to retrieval-based pair selection and parallel global SfM. VGGT-X is faster at smaller scales but GloSplat-F surpasses it at 750+ images due to better asymptotic scaling.}
    \label{fig:runtime}
  \end{center}
  \vskip -0.2in
\end{figure}

\Cref{fig:runtime} presents runtime comparisons on the Courthouse scene from Tanks and Temples, varying the number of input images from 250 to 1000 (see \Cref{tab:runtime} in the appendix for detailed numbers). GloSplat-F demonstrates excellent scalability: while COLMAP's runtime grows super-linearly due to incremental SfM's sequential nature and exhaustive matching, GloSplat-F exhibits near-linear scaling thanks to retrieval-based pair selection and parallel global SfM. At 1000 images, GloSplat-F achieves a \textbf{13.3$\times$ speedup} over COLMAP while delivering superior reconstruction quality (\Cref{tab:comparison}).

Notably, VGGT-X is faster at smaller image counts (250--500 images) due to its feed-forward architecture, but GloSplat-F overtakes it at 750+ images. This crossover occurs because VGGT-X's runtime scales with the number of view pairs processed by its attention mechanism, while GloSplat-F's retrieval limits the matching graph to a constant number of neighbors per image. At 1000 images, GloSplat-F is 1.2$\times$ faster than VGGT-X while achieving significantly higher rendering quality (+1.37 dB PSNR on MipNeRF360). GloSplat-A, using exhaustive matching, is slower but still 3.2$\times$ faster than COLMAP while achieving the highest quality among all methods.

\subsection{GloSplat-A: Surpassing COLMAP-Based Methods}

Our second variant, \textbf{GloSplat-A}, uses exhaustive feature matching to maximize reconstruction quality. \Cref{tab:colmap_comparison} compares against state-of-the-art COLMAP-based 3DGS methods on MipNeRF360.

\begin{table}[t]
  \caption{Comparison with COLMAP-based methods on MipNeRF360. All baselines use COLMAP poses. GloSplat-A achieves SOTA among all methods. Best in \textbf{bold}, second best \underline{underlined}.}
  \label{tab:colmap_comparison}
  \begin{center}
  \resizebox{\columnwidth}{!}{
  \begin{tabular}{lccc}
    \toprule
    Method &PSNR$\uparrow$ &SSIM$\uparrow$ &LPIPS$\downarrow$ \\
    \midrule
    3DGS &27.48 &0.815 &0.216 \\
    AbsGS &27.52 &0.820 &0.198 \\
    PixelGS &27.62 &0.824 &0.189 \\
    MiniSplatting-D &27.57 &0.832 &\underline{0.176} \\
    TamingGS &27.96 &0.822 &0.207 \\
    3DGS-MCMC &28.01 &0.835 &0.186 \\
    SteepGS &27.05 &0.795 &0.247 \\
    Perceptual-GS &27.77 &0.829 &0.187 \\
    Improved-GS &\underline{28.19} &\underline{0.836} &0.186 \\
    \midrule
    \textbf{GloSplat-A (Ours)} &\textbf{28.86} &\textbf{0.862} &\textbf{0.139} \\
    \bottomrule
  \end{tabular}
  }
  \end{center}
\end{table}

GloSplat-A achieves state-of-the-art performance, outperforming all COLMAP-based methods by significant margins. Compared to the previous best method Improved-GS, we improve PSNR by +0.67 dB (+2.4\%), SSIM by 2.6 points (+3.1\%), and reduce LPIPS by 25.3\%. These results validate our core thesis: \emph{joint pose-appearance optimization during 3DGS training, combined with global SfM initialization, provides superior geometric consistency compared to the traditional frozen-pose pipeline}. While COLMAP-based methods treat pose estimation as a preprocessing step with frozen outputs, our joint optimization enables photometric gradients to refine camera poses throughout training, leading to higher-quality reconstructions.

\paragraph{Comparison with 3RGS on ScanNet.}
We additionally compare against 3RGS~\cite{huang20253r} on ScanNet~\cite{dai2017scannet} to evaluate both pose estimation accuracy and rendering quality. \Cref{tab:scannet} reports rotation error (R), absolute trajectory error (ATE), and PSNR. GloSplat-F achieves the best pose accuracy on all scenes while also delivering higher rendering quality, demonstrating that our joint photometric-geometric optimization provides superior pose refinement compared to 3RGS's photometric-only approach.

\begin{table}[t]
    \centering
    \caption{Comparison with COLMAP and 3RGS on ScanNet. Best in \textbf{bold}, second best \underline{underlined}.}
    \label{tab:scannet}
    \resizebox{\columnwidth}{!}{
    \begin{tabular}{lccc|ccc|ccc}
    \toprule
    & \multicolumn{3}{c|}{COLMAP} & \multicolumn{3}{c|}{3RGS} & \multicolumn{3}{c}{\textbf{GloSplat-F (Ours)}} \\
    \cmidrule(r){2-4} \cmidrule(r){5-7} \cmidrule(r){8-10}
    Scene & R(°)$\downarrow$ & ATE(m)$\downarrow$ & PSNR$\uparrow$ & R(°)$\downarrow$ & ATE(m)$\downarrow$ & PSNR$\uparrow$ & R(°)$\downarrow$ & ATE(m)$\downarrow$ & PSNR$\uparrow$ \\
    \midrule
    0079\_00 & 3.55 & \underline{0.014} & 30.78 & \underline{2.45} & \underline{0.014} & \underline{32.58} & \textbf{2.12} & \textbf{0.011} & \textbf{33.42} \\
    0301\_00 & 133.83 & 0.169 & 23.63 & \underline{9.30} & \underline{0.009} & \underline{30.11} & \textbf{7.82} & \textbf{0.007} & \textbf{31.24} \\
    0418\_00 & 5.03 & \underline{0.012} & 29.03 & \underline{4.34} & \underline{0.012} & \underline{31.62} & \textbf{3.78} & \textbf{0.010} & \textbf{32.35} \\
    \bottomrule
    \end{tabular}
    }
\end{table}

Qualitative comparisons are presented in \Cref{fig:qualitative} in the appendix.

\section{Conclusion}
\label{sec:conclusion}

We have presented GloSplat, a unified framework for 3D reconstruction that introduces a key architectural novelty: preserving explicit SfM feature tracks as first-class entities during 3D Gaussian Splatting training, with track 3D points maintained as separate optimizable parameters from Gaussian means. Unlike prior joint optimization methods (BARF, NeRF--, 3RGS) that rely purely on photometric gradients and suffer from early-stage pose drift, our dual photometric-geometric supervision provides persistent anchoring that stabilizes optimization while enabling fine-grained pose refinement. Our two variants, GloSplat-F (retrieval-based) and GloSplat-A (exhaustive matching), achieve state-of-the-art results in their respective categories: GloSplat-F establishes new performance standards among COLMAP-free methods while offering significant speedups, and GloSplat-A surpasses all COLMAP-based baselines, demonstrating that joint pose-appearance optimization with global SfM can outperform incremental approaches.

\paragraph{Limitations and Future Work.}
Our work has several limitations that warrant future investigation. First, the feature extraction, matching, and pair selection stages remain frozen preprocessing---gradients do not flow back through these components. Consequently, our joint optimization operates only during the 3DGS training phase, refining camera poses and Gaussian primitives but not the upstream modules. The performance gap between GloSplat-F (27.77 dB) and GloSplat-A (28.86 dB) on MipNeRF360---a difference of 1.09 dB---reflects both the sparser matching graph from retrieval-based selection and differences between learned (XFeat) and classical (SIFT) features. When retrieval-based pair selection misses important overlapping views, global SfM initialization degrades, and subsequent joint optimization cannot fully recover. Future work should explore more robust matching strategies or learned retrieval methods that better identify challenging but informative image pairs.

Second, a fully end-to-end differentiable approach, where gradients from rendering losses flow back through SfM and into the feature extractor itself, remains an open challenge. Such a unified architecture would enable the feature network to learn representations optimized for downstream reconstruction quality rather than generic matching performance, though this requires significant engineering effort and may introduce stability challenges during training.

\section*{Impact Statement}

This paper presents work whose goal is to advance the field of 3D reconstruction and novel view synthesis. We hope that GloSplat's success in jointly optimizing camera poses and radiance fields will inspire the broader computer vision community to reconsider frozen-interface designs between pipeline stages. The demonstrated gains from joint pose-appearance optimization---where camera poses continue to receive gradients during appearance learning---suggest that similar principles may benefit other multi-stage vision pipelines, from SLAM systems to multi-modal reconstruction. We encourage researchers to consider cross-stage gradient flow when designing future 3D vision systems, as the boundaries between ``preprocessing'' and ``main tasks'' are often artificial constraints inherited from historical software architectures rather than fundamental algorithmic necessities.

\newpage

\bibliography{main}
\bibliographystyle{icml2026}

\newpage
\appendix
\onecolumn

\section{Qualitative Results}
\label{sec:qualitative_appendix}

\begin{figure*}[h]
  \begin{center}
    \begin{subfigure}[b]{\textwidth}
      \centering
      \includegraphics[width=0.19\linewidth]{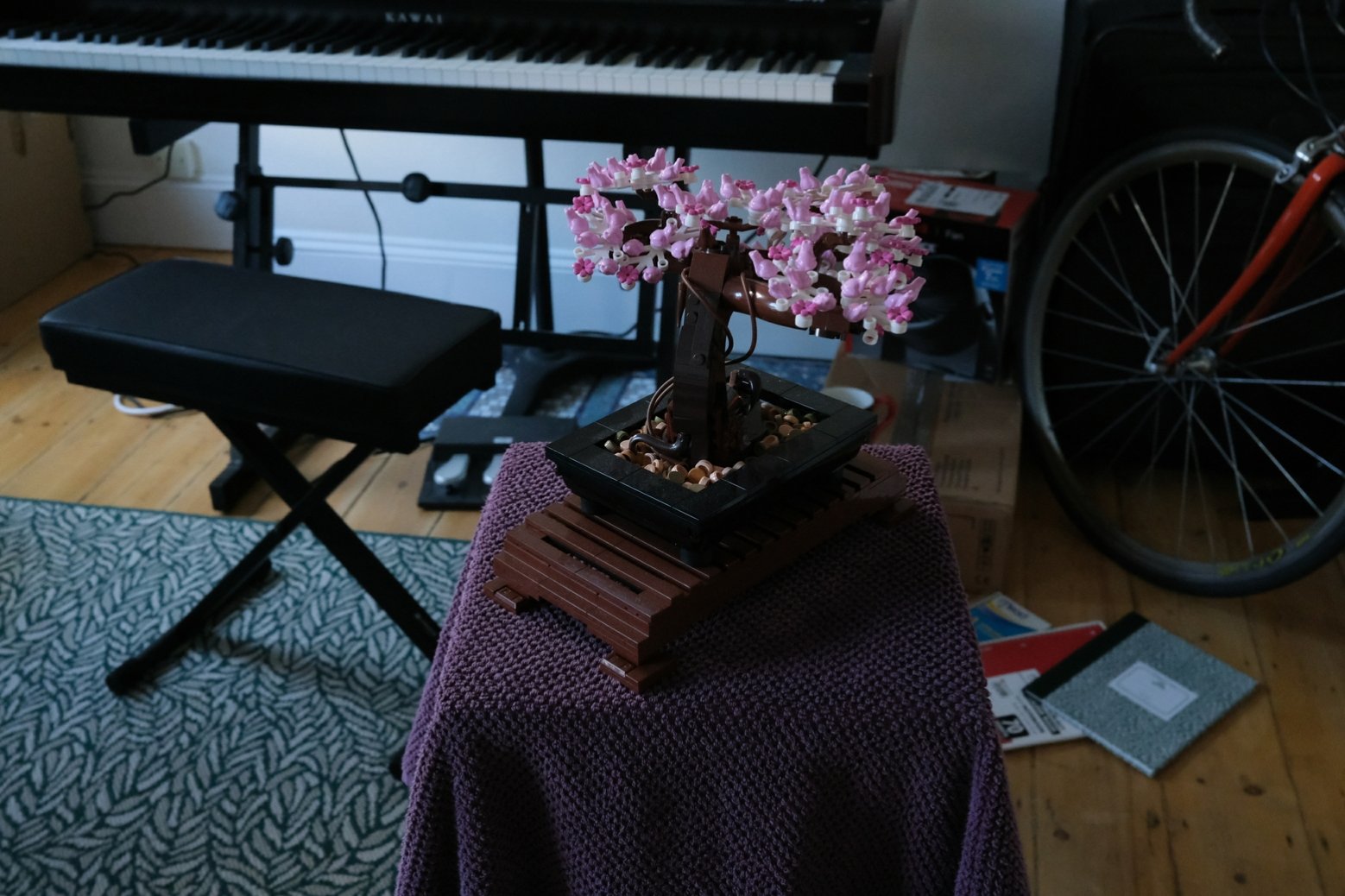}
      \includegraphics[width=0.19\linewidth]{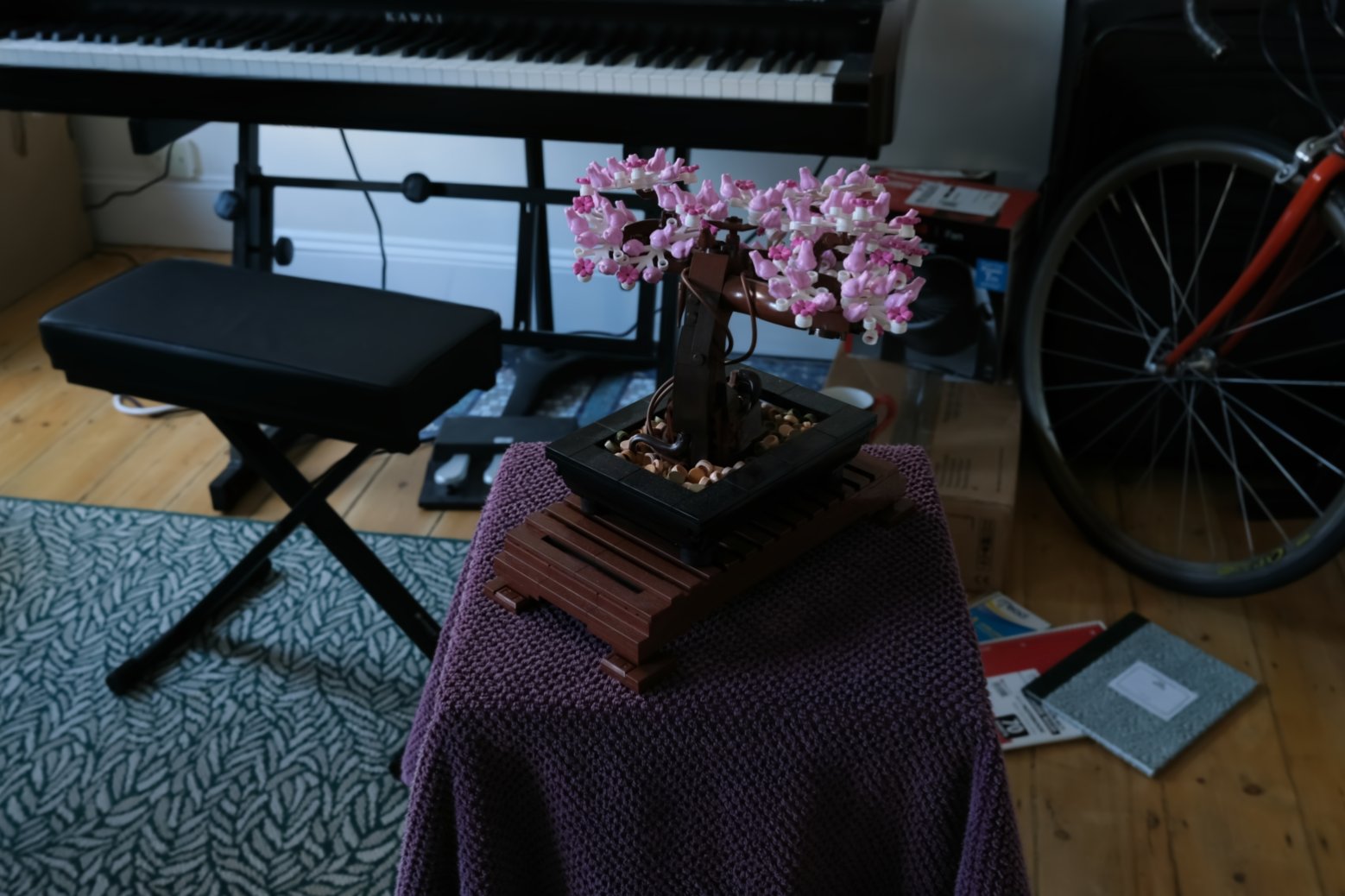}
      \includegraphics[width=0.19\linewidth]{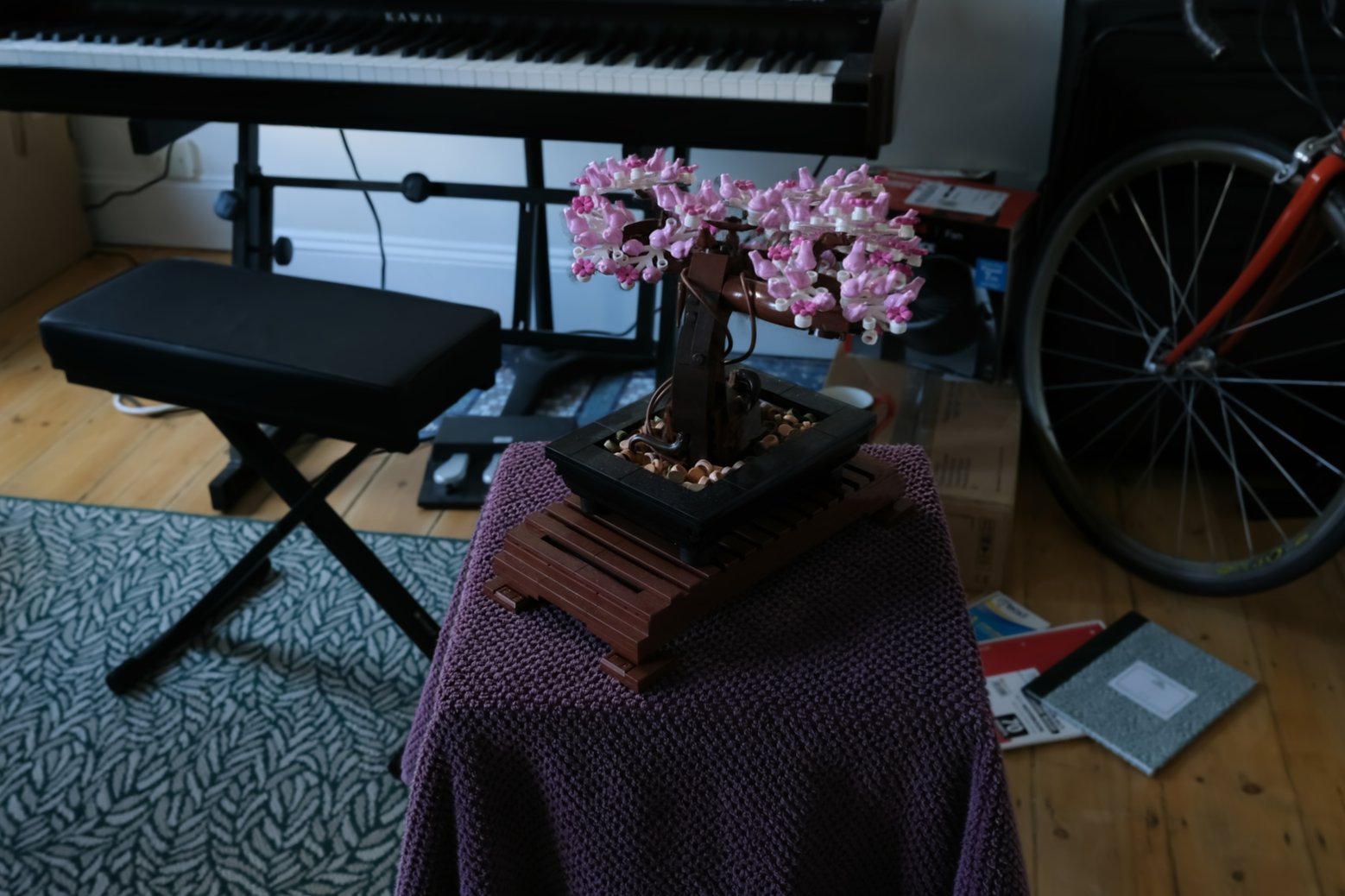}
      \includegraphics[width=0.19\linewidth]{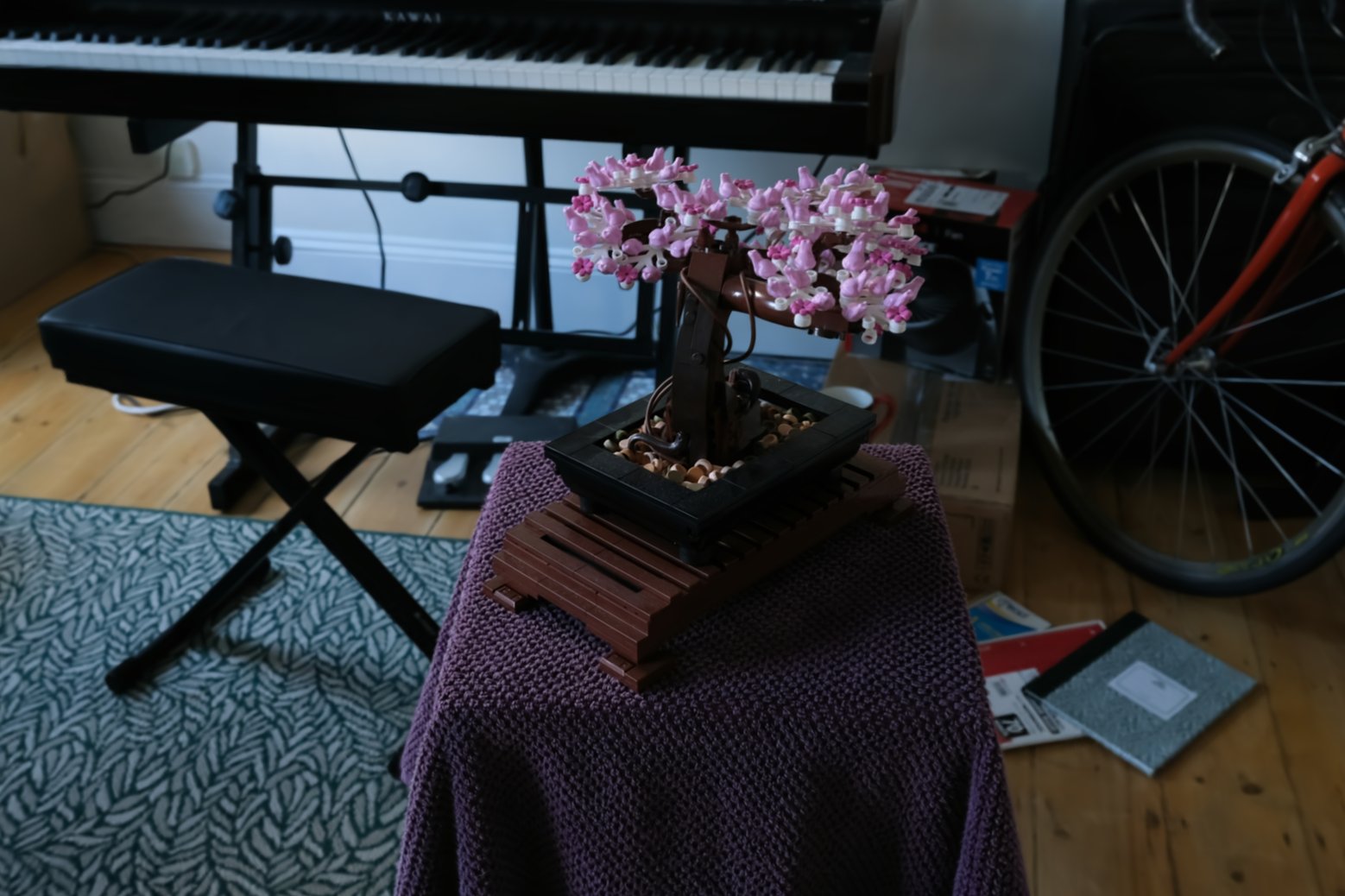}
      \includegraphics[width=0.19\linewidth]{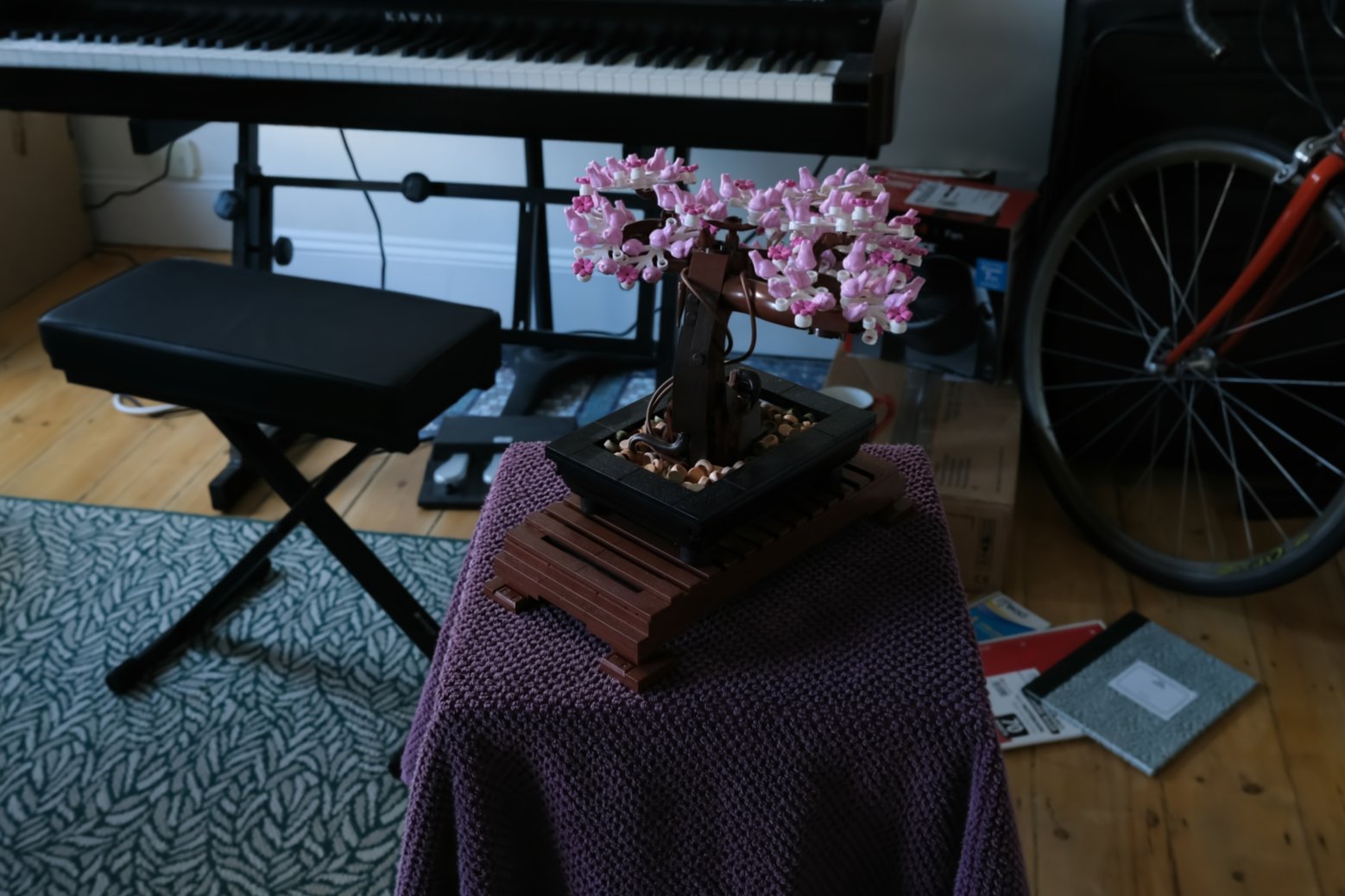}
      \vspace{-2mm}
      \caption*{\textbf{Bonsai} --- GloSplat-A: 36.39 dB / 0.060 LPIPS, VGGT-X: 27.82 dB / 0.083 LPIPS}
    \end{subfigure}
    \vspace{1mm}
    \begin{subfigure}[b]{\textwidth}
      \centering
      \includegraphics[width=0.19\linewidth]{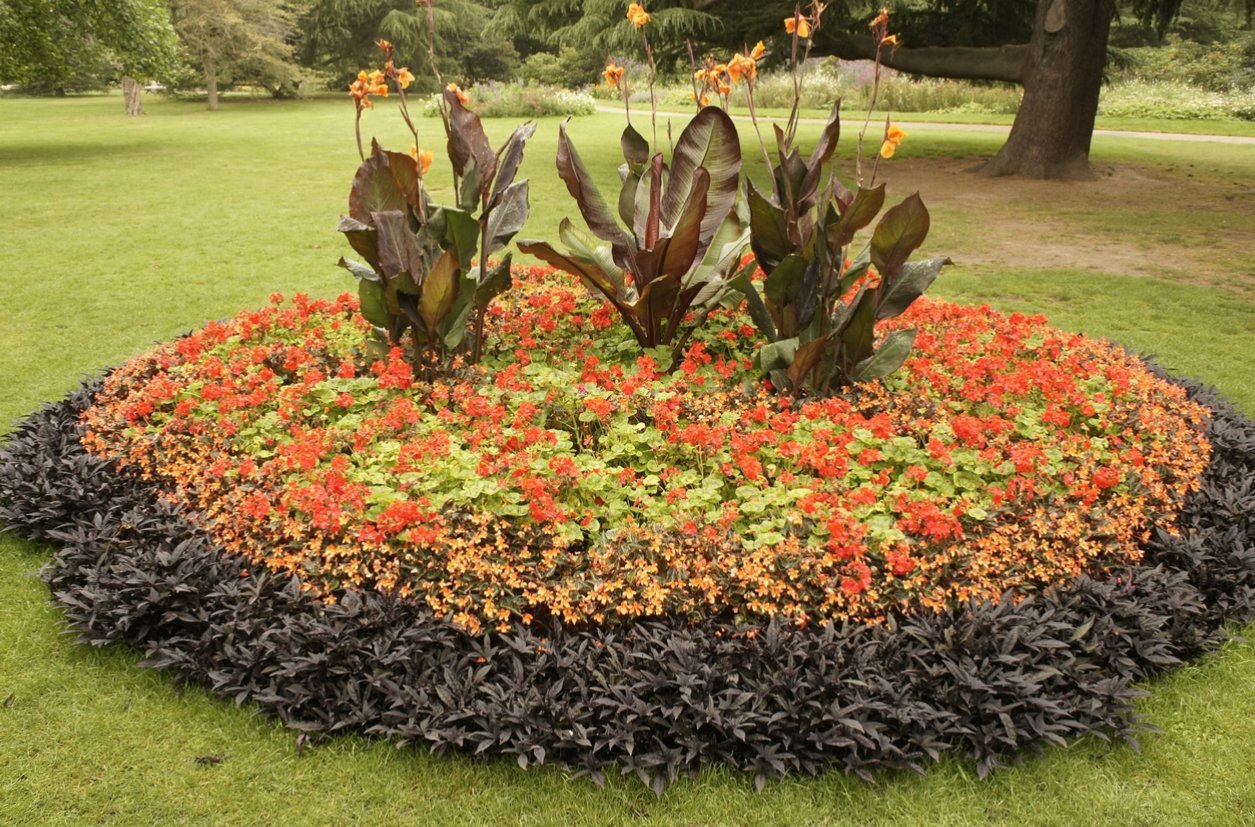}
      \includegraphics[width=0.19\linewidth]{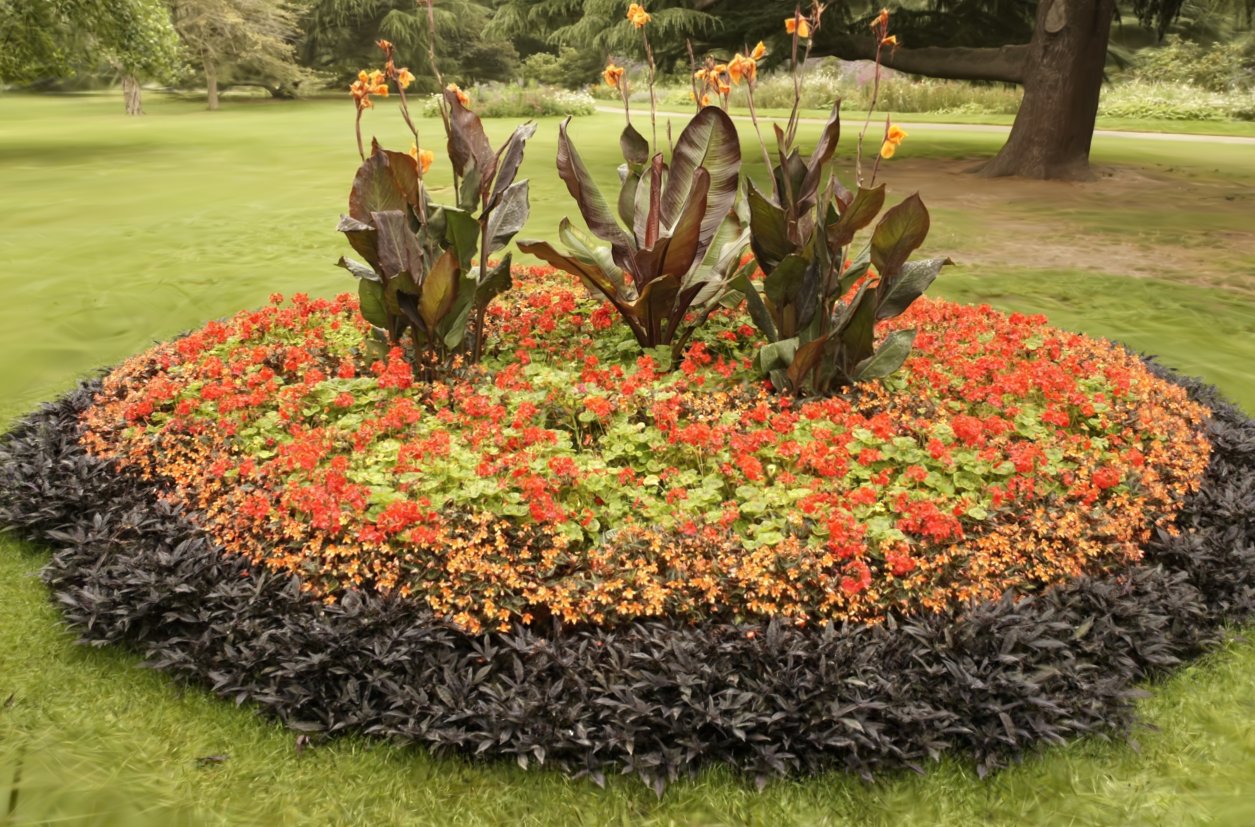}
      \includegraphics[width=0.19\linewidth]{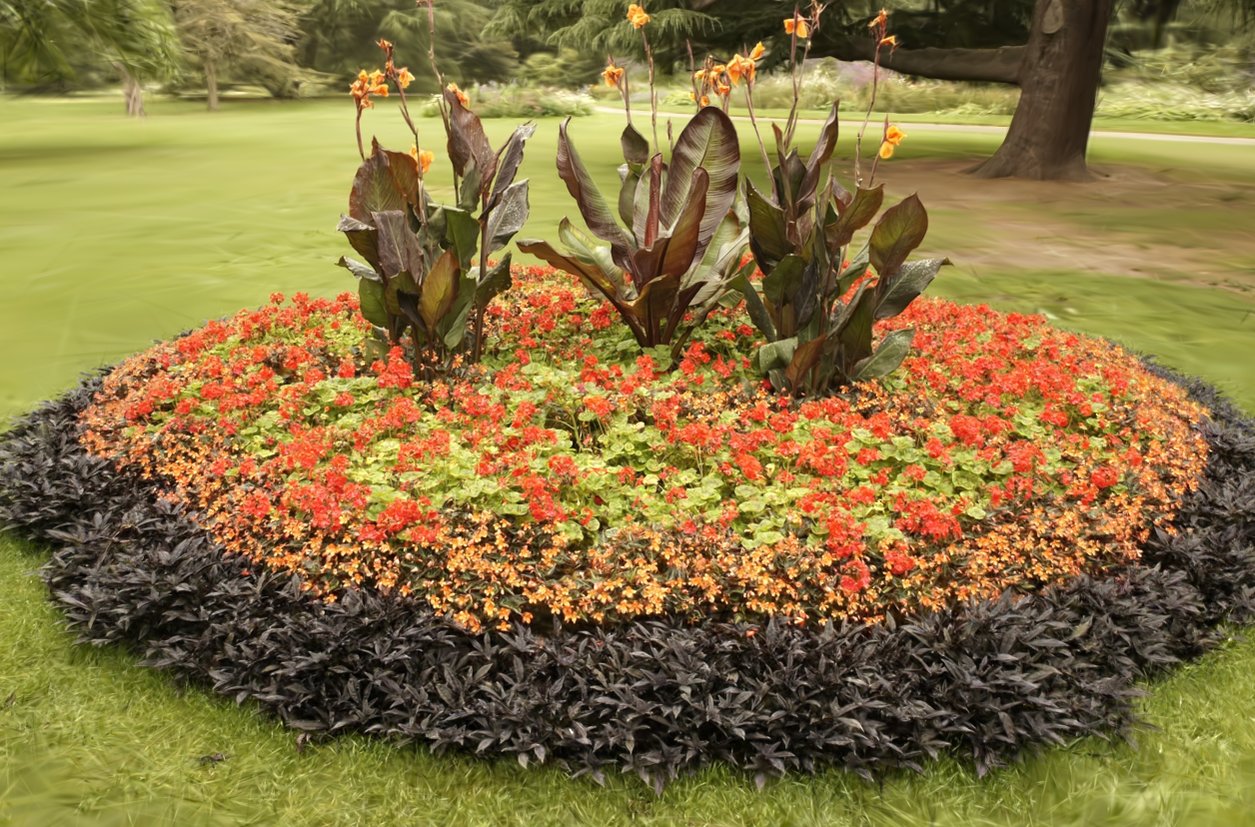}
      \includegraphics[width=0.19\linewidth]{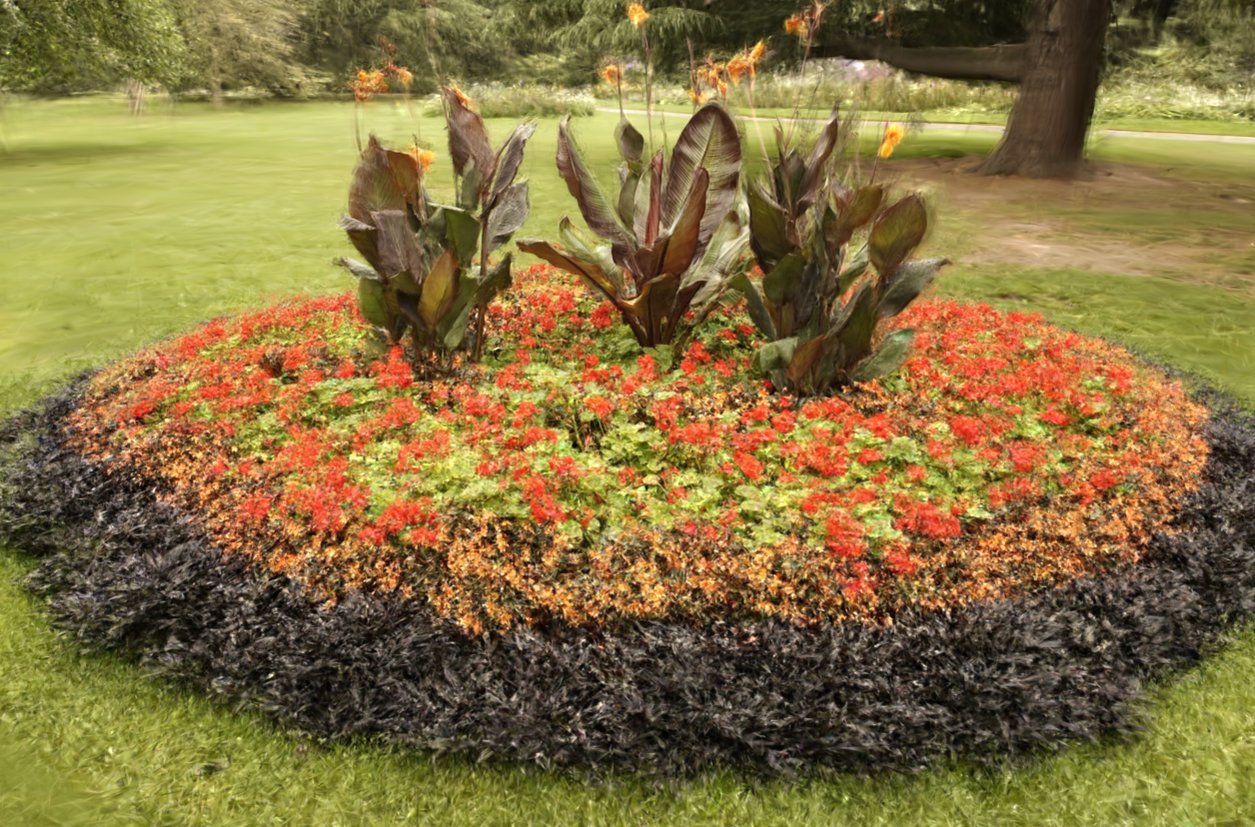}
      \includegraphics[width=0.19\linewidth]{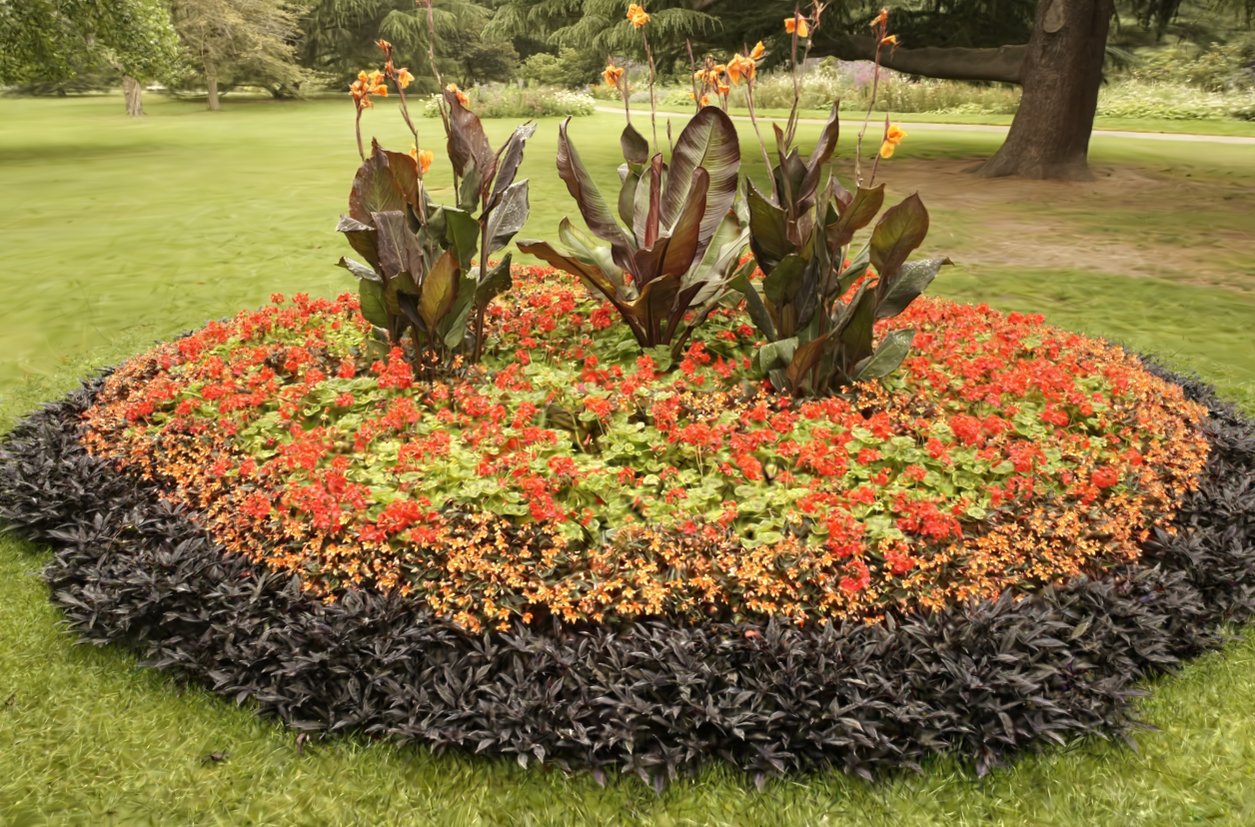}
      \vspace{-2mm}
      \caption*{\textbf{Flowers} --- GloSplat-A: 24.88 dB / 0.147 LPIPS, VGGT-X: 18.13 dB / 0.273 LPIPS}
    \end{subfigure}
    \vspace{1mm}
    \begin{subfigure}[b]{\textwidth}
      \centering
      \includegraphics[width=0.19\linewidth]{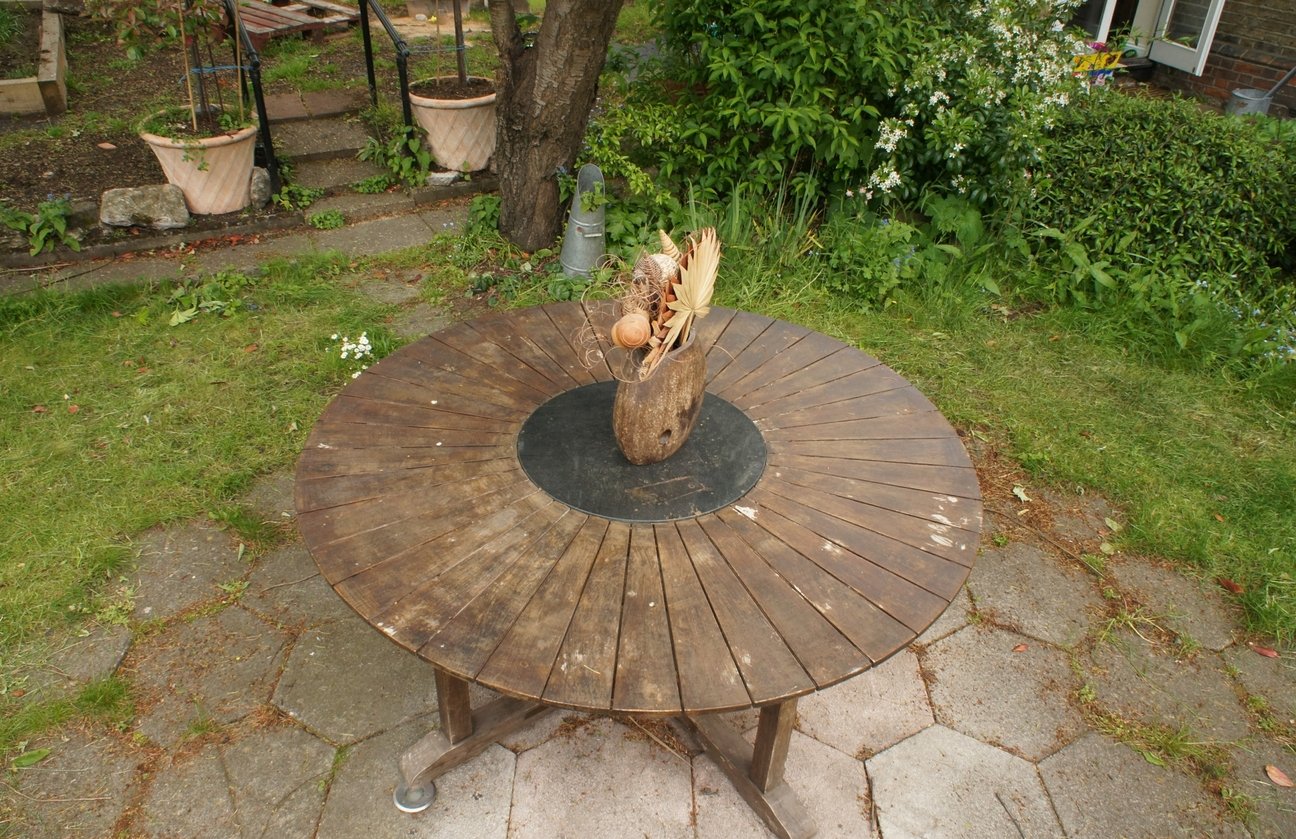}
      \includegraphics[width=0.19\linewidth]{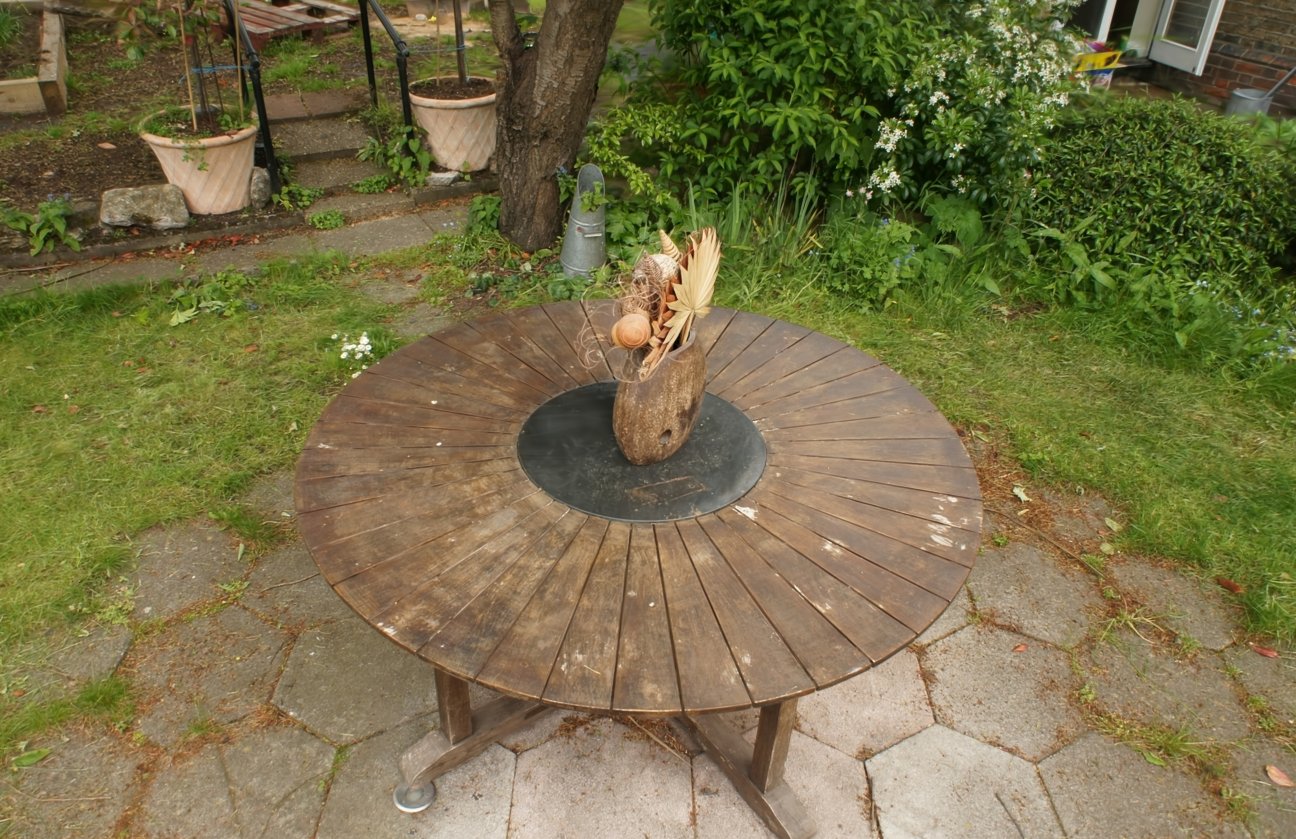}
      \includegraphics[width=0.19\linewidth]{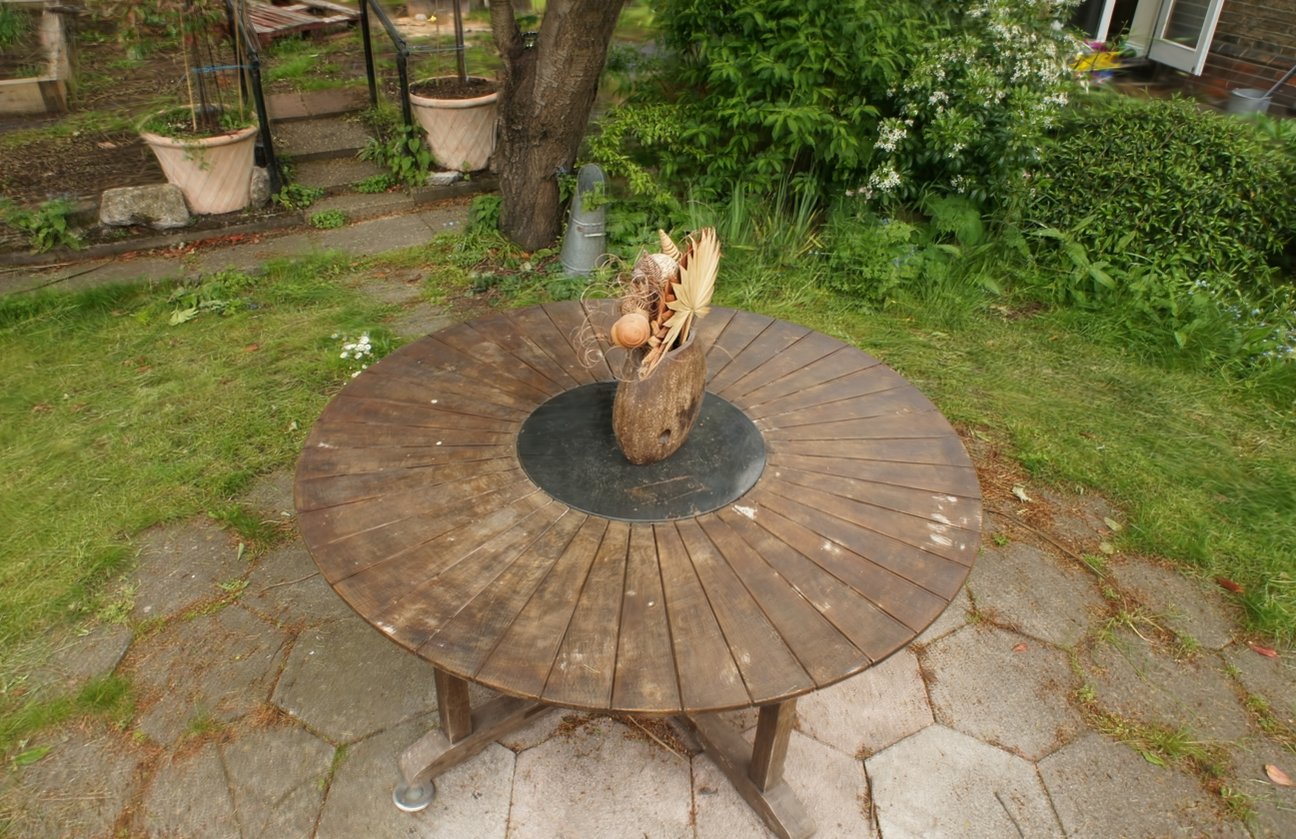}
      \includegraphics[width=0.19\linewidth]{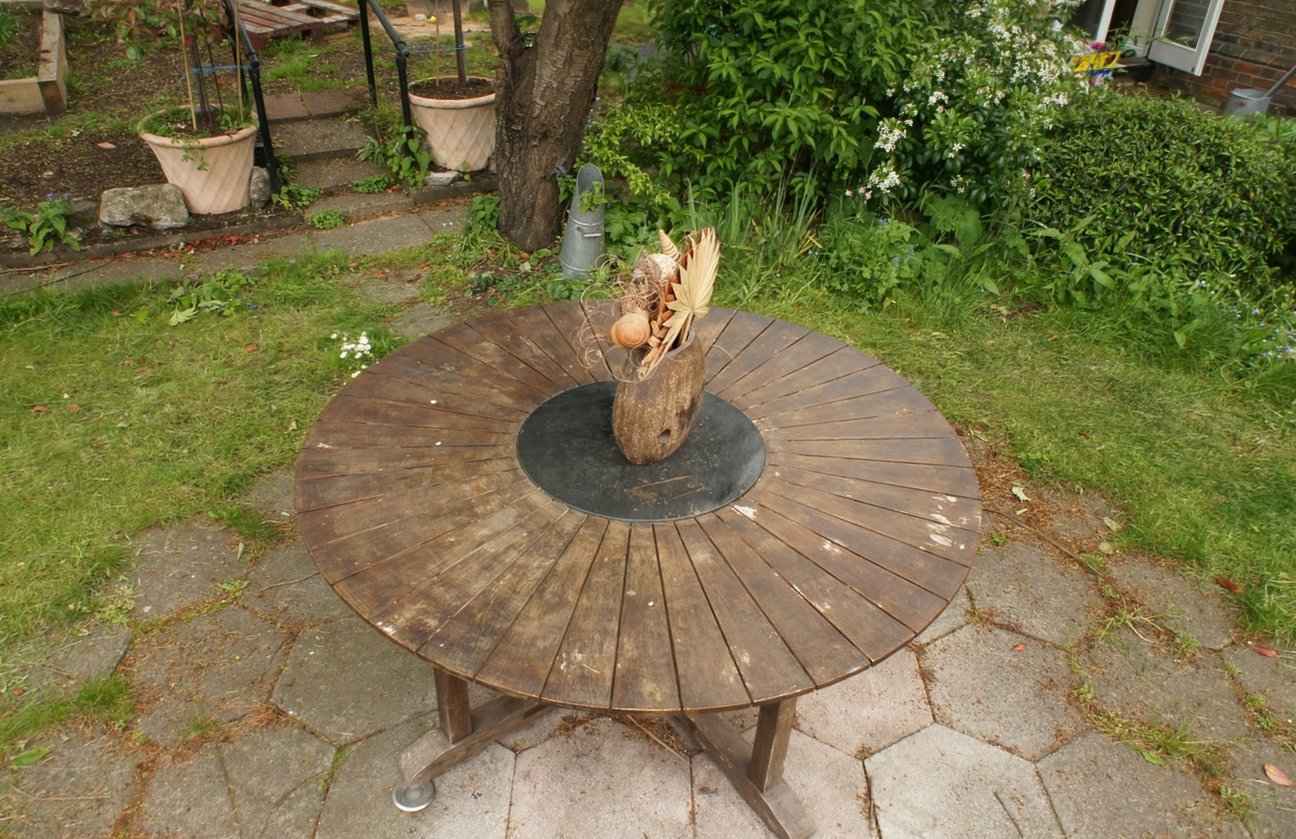}
      \includegraphics[width=0.19\linewidth]{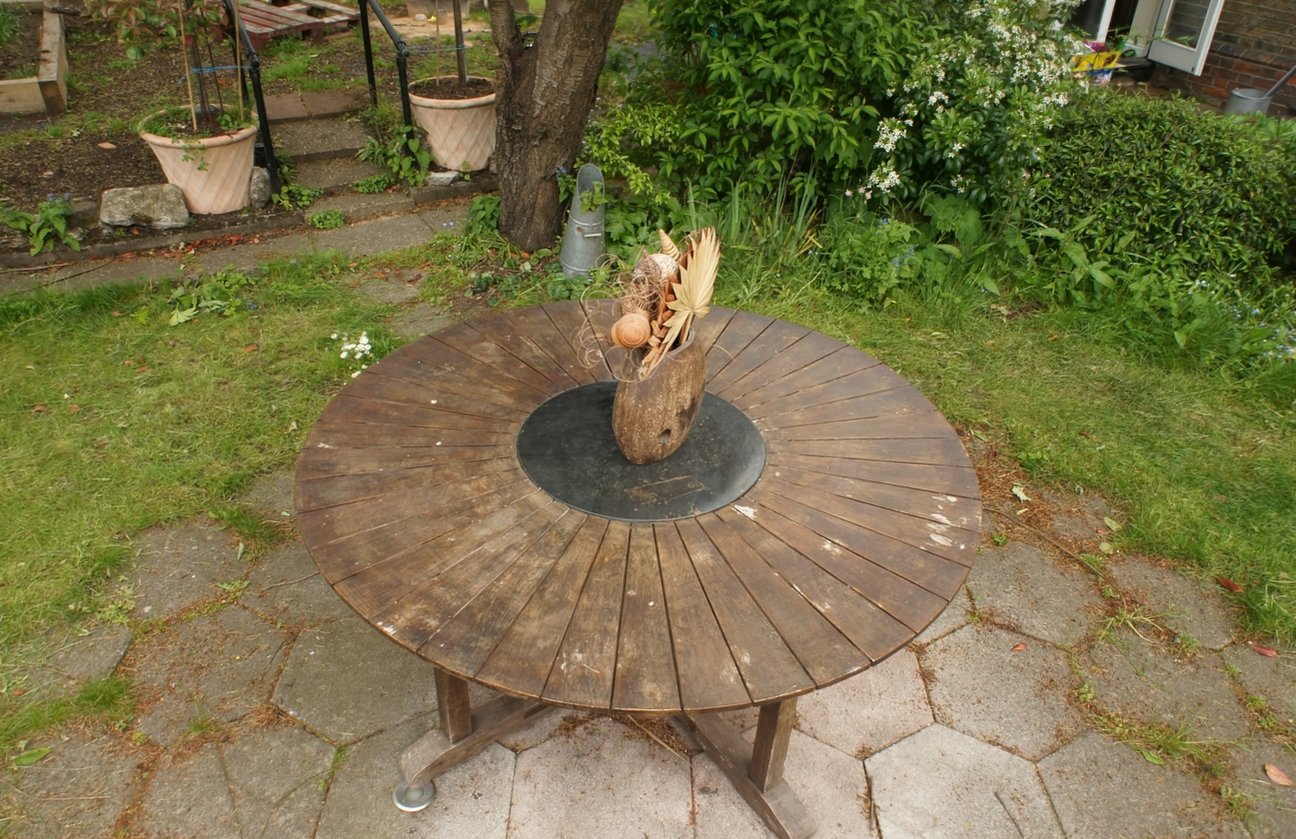}
      \vspace{-2mm}
      \caption*{\textbf{Garden} --- GloSplat-A: 30.82 dB / 0.074 LPIPS, VGGT-X: 25.13 dB / 0.101 LPIPS}
    \end{subfigure}
    \vspace{1mm}
    \begin{subfigure}[b]{\textwidth}
      \centering
      \makebox[0.19\linewidth]{\small GT}
      \makebox[0.19\linewidth]{\small GloSplat-A (Ours)}
      \makebox[0.19\linewidth]{\small GloSplat-F (Ours)}
      \makebox[0.19\linewidth]{\small VGGT-X}
      \makebox[0.19\linewidth]{\small Improved-GS}
    \end{subfigure}
    \caption{\textbf{Qualitative Comparison on MipNeRF360.} We compare novel view synthesis results from GloSplat-A, GloSplat-F, VGGT-X, and Improved-GS against ground truth. GloSplat-A achieves significantly higher PSNR and lower LPIPS across all scenes. On \textit{Bonsai}, GloSplat-A outperforms VGGT-X by \textbf{+8.57 dB PSNR}. On \textit{Flowers}, our method shows \textbf{+6.76 dB PSNR} and \textbf{46\% lower LPIPS} (0.147 vs 0.273) over VGGT-X. On \textit{Garden}, GloSplat-A achieves \textbf{+5.69 dB} over VGGT-X. Best viewed zoomed in.}
    \label{fig:qualitative}
  \end{center}
\end{figure*}

\Cref{fig:qualitative} presents visual comparisons on representative scenes from MipNeRF360. On the \textit{Bonsai} scene, GloSplat-A produces notably sharper leaf structures and intricate branch details, achieving 36.39 dB PSNR compared to VGGT-X's 27.82 dB---a remarkable 8.57 dB improvement. The \textit{Flowers} scene demonstrates GloSplat's ability to handle challenging thin structures: our method achieves 46\% better perceptual quality (0.147 vs 0.273 LPIPS) with cleaner flower petal boundaries. On the outdoor \textit{Garden} scene, joint pose-appearance optimization produces cleaner foliage edges and more accurate colors, outperforming VGGT-X by 5.69 dB. These results illustrate that continuous pose refinement during 3DGS training is particularly beneficial for scenes with fine details and complex structures.

\section{Per-Scene Results}
\label{sec:perscene}

We report detailed per-scene results for GloSplat-F on all three benchmark datasets. These tables complement the aggregate results in \Cref{tab:comparison} by revealing performance variation across different scene characteristics. Notably, GloSplat-F achieves particularly strong results on object-centric scenes (Apple, Skateboard, Teddybear in CO3Dv2 with PSNR $>$ 33 dB), while maintaining competitive performance on challenging outdoor scenes with complex geometry and lighting (Flowers, Stump, Treehill in MipNeRF360). The consistent performance across diverse scene types---from indoor tabletop captures to large-scale outdoor environments---validates that our joint pose-appearance optimization generalizes well without scene-specific tuning.

\begin{table}[h]
  \caption{Per-scene results on CO3Dv2 dataset.}
  \label{tab:co3d_perscene}
  \begin{center}
  \begin{tabular}{lccc}
    \toprule
    Scene &PSNR$\uparrow$ &SSIM$\uparrow$ &LPIPS$\downarrow$ \\
    \midrule
    Apple &35.30 &0.952 &0.063 \\
    Bench &32.33 &0.931 &0.089 \\
    Hydrant &29.33 &0.920 &0.071 \\
    Skateboard &33.35 &0.943 &0.116 \\
    Teddybear &33.24 &0.932 &0.102 \\
    \midrule
    Average &32.71 &0.936 &0.088 \\
    \bottomrule
  \end{tabular}
  \end{center}
\end{table}

\begin{table}[h]
  \caption{Per-scene results on MipNeRF360 dataset.}
  \label{tab:mip360_perscene}
  \begin{center}
  \begin{tabular}{lccc}
    \toprule
    Scene &PSNR$\uparrow$ &SSIM$\uparrow$ &LPIPS$\downarrow$ \\
    \midrule
    Bicycle &25.98 &0.802 &0.145 \\
    Bonsai &33.16 &0.952 &0.114 \\
    Counter &29.60 &0.921 &0.131 \\
    Flowers &22.93 &0.688 &0.244 \\
    Garden &27.64 &0.866 &0.084 \\
    Kitchen &32.99 &0.942 &0.085 \\
    Room &30.98 &0.917 &0.154 \\
    Stump &24.27 &0.650 &0.209 \\
    Treehill &22.34 &0.621 &0.308 \\
    \midrule
    Average &27.77 &0.818 &0.164 \\
    \bottomrule
  \end{tabular}
  \end{center}
\end{table}

\begin{table}[h]
  \caption{Per-scene results on Tanks and Temples dataset.}
  \label{tab:tnt_perscene}
  \begin{center}
  \begin{tabular}{lccc}
    \toprule
    Scene &PSNR$\uparrow$ &SSIM$\uparrow$ &LPIPS$\downarrow$ \\
    \midrule
    Barn &26.43 &0.842 &0.168 \\
    Caterpillar &24.89 &0.851 &0.162 \\
    Ignatius &27.15 &0.912 &0.121 \\
    Truck &25.31 &0.867 &0.148 \\
    Train &25.32 &0.873 &0.156 \\
    \midrule
    Average &25.82 &0.869 &0.151 \\
    \bottomrule
  \end{tabular}
  \end{center}
\end{table}

\section{Runtime Comparison}
\label{sec:runtime_appendix}

We provide detailed runtime measurements for full-pipeline reconstruction on the Courthouse scene from Tanks and Temples, which contains up to 1106 images and serves as a representative benchmark for scalability analysis. All experiments were conducted on a server with an NVIDIA RTX PRO 6000 GPU (96GB memory), AMD Ryzen 9 9950X CPU, running Ubuntu 24.04. \textbf{For fair comparison, COLMAP was compiled with CUDA support and run with GPU acceleration enabled for all applicable stages} (feature extraction via SIFT-GPU, exhaustive matching, and geometric verification). Runtime includes all stages: feature extraction, pair selection (retrieval or exhaustive), feature matching, SfM (incremental for COLMAP, global for GloSplat), and 3DGS training (30k iterations). As shown in \Cref{tab:runtime}, GloSplat-F demonstrates superior asymptotic scaling due to its $O(n)$ retrieval-based pair selection and parallel global SfM, achieving a 13.3$\times$ speedup over COLMAP at 1000 images. The crossover point with VGGT-X occurs around 750 images, after which GloSplat-F's linear scaling provides increasing advantages.

\begin{table}[h]
  \caption{Runtime comparison (seconds) on Courthouse scene with varying image counts. All methods run on the same hardware (RTX PRO 6000 GPU); COLMAP uses GPU-accelerated SIFT and matching. GloSplat-F achieves superior scaling due to $O(n)$ pair selection and parallel global SfM.}
  \label{tab:runtime}
  \begin{center}
  \begin{tabular}{lcccc}
    \toprule
    Method &250 imgs &500 imgs &750 imgs &1000 imgs \\
    \midrule
    COLMAP &644.6 &1165.0 &2226.3 &4349.2 \\
    3RGS (MASt3R) &312.2 &638.0 &978.5 &1308.2 \\
    GloSplat-A &222.2 &408.2 &855.7 &1371.3 \\
    VGGT-X &53.9 &113.3 &249.6 &398.9 \\
    \textbf{GloSplat-F} &100.2 &191.9 &250.0 &327.8 \\
    \midrule
    \textit{Speedup vs COLMAP} & & & & \\
    \quad GloSplat-F &6.4$\times$ &6.1$\times$ &8.9$\times$ &13.3$\times$ \\
    \bottomrule
  \end{tabular}
  \end{center}
\end{table}

\paragraph{Training Overhead of Joint Optimization.}
We additionally measure the training time overhead introduced by our joint BA loss compared to vanilla MCMC 3DGS training. \Cref{tab:training_overhead} reports training times on a representative scene with 200 images. The joint optimization adds only marginal overhead ($\sim$3\%) over vanilla MCMC 3DGS, as the reprojection loss computation on sparse feature tracks is computationally lightweight compared to the differentiable rasterization that dominates training time. This demonstrates that our geometric anchoring comes at negligible computational cost while providing significant quality improvements (+0.81 dB as shown in \Cref{tab:extended_ablation}).

\begin{table}[h]
  \caption{Training time comparison between vanilla MCMC 3DGS and MCMC 3DGS with joint optimization (30k iterations, 200 images). Joint optimization adds minimal overhead while providing +0.81 dB improvement.}
  \label{tab:training_overhead}
  \begin{center}
  \begin{tabular}{lcc}
    \toprule
    Configuration &Training Time &Overhead \\
    \midrule
    MCMC 3DGS (vanilla) &$\sim$31 min &-- \\
    MCMC 3DGS + Joint Opt (Ours) &$\sim$32 min &+3\% \\
    \bottomrule
  \end{tabular}
  \end{center}
\end{table}

\section{Extended Ablation Study}
\label{sec:extended_ablation}

We conduct comprehensive ablation experiments on MipNeRF360 to isolate the contribution of each pipeline component. \Cref{tab:extended_ablation} presents results organized by component category, with detailed analysis below.

\begin{table}[h]
  \caption{Ablation study on MipNeRF360. We show reference configurations (top) and ablations organized by component. $\Delta$ shows change from GloSplat-F. Best in \textbf{bold}; ``--'' indicates failure.}
  \label{tab:extended_ablation}
  \begin{center}
  \begin{tabular}{lcccc}
    \toprule
    Configuration &PSNR$\uparrow$ &$\Delta$ &SSIM$\uparrow$ &LPIPS$\downarrow$ \\
    \midrule
    \multicolumn{5}{l}{\textit{Reference Configurations}} \\
    \textbf{GloSplat-A} (exhaustive matching) &\textbf{28.86} &\textbf{+1.09} &\textbf{0.862} &\textbf{0.139} \\
    \textbf{GloSplat-F} (retrieval matching) &27.77 &-- &0.818 &0.164 \\
    MCMC$^\dagger$ (COLMAP, frozen poses) &27.91 &+0.14 &0.836 &0.154 \\
    \midrule
    \multicolumn{5}{l}{\textit{(A) Joint Optimization --- removing from GloSplat-F}} \\
    \quad Frozen poses after Global SfM &19.18 &--8.59 &0.463 &0.472 \\
    \quad Photometric-only (no BA loss) &26.96 &--0.81 &0.783 &0.190 \\
    \midrule
    \multicolumn{5}{l}{\textit{(B) SfM Backend --- isolating Global SfM vs COLMAP}} \\
    \quad COLMAP (exhaustive) + Joint Opt &28.52 &+0.75 &0.854 &0.144 \\
    \quad COLMAP (retrieval) + no Joint Opt &17.93 &--9.84 &0.445 &0.528 \\
    \midrule
    \multicolumn{5}{l}{\textit{(C) Densification Strategy}} \\
    \quad Standard densification (no MCMC) &26.02 &--1.75 &0.746 &0.239 \\
    \midrule
    \multicolumn{5}{l}{\textit{(D) Feature Matching Pipeline}} \\
    \quad SuperPoint + SuperGlue &27.06 &--0.71 &0.791 &0.187 \\
    \quad DISK + LightGlue &19.65 &--8.12 &0.494 &0.516 \\
    \quad R2D2 + NN Matcher &17.54 &--10.23 &0.473 &0.673 \\
    \midrule
    \multicolumn{5}{l}{\textit{(E) Image Retrieval}} \\
    \quad NetVLAD &26.35 &--1.42 &0.762 &0.208 \\
    \quad OpenIBL &20.68 &--7.09 &0.554 &0.452 \\
    \quad DIR (AP-GeM) &-- &-- &-- &-- \\
    \bottomrule
  \end{tabular}
  \end{center}
\end{table}

\paragraph{Reference Configurations.}
We include three reference points: \textbf{GloSplat-A} (28.86 dB) uses exhaustive matching for maximum quality; \textbf{GloSplat-F} (27.77 dB) uses retrieval-based matching for speed; \textbf{MCMC$^\dagger$} (27.91 dB) is the COLMAP baseline with frozen poses. The $\Delta$ column shows change relative to GloSplat-F.

\paragraph{(A) Joint Optimization.}
We ablate the joint pose-appearance optimization from GloSplat-F:
\begin{itemize}
    \item \textbf{Frozen poses}: Disabling all pose optimization after Global SfM causes --8.59 dB degradation, showing that pose refinement during 3DGS training is essential.
    \item \textbf{Photometric-only}: Removing the BA loss (poses receive only photometric gradients) causes --0.81 dB degradation, isolating the geometric anchoring contribution.
\end{itemize}

\paragraph{(B) SfM Backend.}
To isolate global SfM's contribution from joint optimization:
\begin{itemize}
    \item \textbf{COLMAP (exhaustive) + Joint Opt}: Using COLMAP initialization with our joint BA loss achieves 28.52 dB. Compared to MCMC$^\dagger$ (27.91 dB, frozen), joint optimization adds \textbf{+0.61 dB}. Compared to GloSplat-A (28.86 dB), global SfM adds \textbf{+0.34 dB}. This confirms: \emph{joint optimization is the primary contributor (+0.61 dB), while global SfM provides additional gains through better initialization (+0.34 dB)}.
    \item \textbf{COLMAP (retrieval)}: COLMAP with sparse retrieval-based matching fails catastrophically (--9.84 dB), as incremental SfM cannot handle sparse view graphs.
\end{itemize}

\paragraph{(C) Densification Strategy.}
MCMC densification~\cite{mcmc3dgs} contributes +1.75 dB. We transparently acknowledge this is an adopted component, not a novel contribution.

\paragraph{(D) Feature Matching (GloSplat-F).}
SuperPoint+SuperGlue achieves competitive results (--0.71 dB). DISK and R2D2 cause severe degradation, indicating XFeat+LightGlue is well-suited for the fast variant.

\paragraph{(E) Image Retrieval.}
NetVLAD shows --1.42 dB vs MegaLoc. OpenIBL/DIR fail due to domain mismatch.

\paragraph{Attribution Summary.}
We decompose GloSplat-A's advantage over MCMC$^\dagger$ (+0.95 dB):
\begin{center}
\begin{tabular}{lcc}
\textbf{Component} & \textbf{Contribution} & \textbf{Fraction} \\
\hline
Joint optimization & +0.61 dB & 64\% \\
Global SfM & +0.34 dB & 36\% \\
\end{tabular}
\end{center}
GloSplat is a \emph{systems} contribution where joint optimization is the primary driver (+0.61 dB, 64\%), validating our core thesis that continuous pose refinement during 3DGS training improves reconstruction quality. Global SfM provides additional gains through better initialization (+0.34 dB, 36\%), and MCMC densification (+1.75 dB) is an adopted component. The full system's performance depends on careful integration of all components.

\paragraph{Why Separate Track Points from Gaussian Means?}
A natural question is whether maintaining track 3D points $\{\mathbf{X}_k\}$ as \emph{separate} parameters from Gaussian means $\{\boldsymbol{\mu}_j\}$ is necessary, or whether one could simply merge them---using Gaussian positions directly for both rendering and reprojection constraints. We ablate this design choice by initializing Gaussians at track point locations and applying the reprojection loss directly to the corresponding Gaussian means. \Cref{tab:track_separation} reports results on MipNeRF360.

\begin{table}[h]
  \caption{Ablation: Separating track points from Gaussian means. Merging tracks with Gaussians degrades all metrics for both variants, validating our architectural choice.}
  \label{tab:track_separation}
  \begin{center}
  \begin{tabular}{lccc}
    \toprule
    Configuration &PSNR$\uparrow$ &SSIM$\uparrow$ &LPIPS$\downarrow$ \\
    \midrule
    GloSplat-A (separate tracks) &\textbf{28.86} &\textbf{0.862} &\textbf{0.139} \\
    GloSplat-A (merged tracks) &28.64 &0.858 &0.144 \\
    \quad $\Delta$ &--0.22 &--0.004 &+0.005 \\
    \midrule
    GloSplat-F (separate tracks) &\textbf{27.77} &\textbf{0.818} &\textbf{0.164} \\
    GloSplat-F (merged tracks) &27.58 &0.813 &0.168 \\
    \quad $\Delta$ &--0.19 &--0.005 &+0.004 \\
    \bottomrule
  \end{tabular}
  \end{center}
\end{table}

Naively merging track points with Gaussian means causes consistent degradation across all metrics. GloSplat-A degrades by --0.22 dB PSNR, while GloSplat-F shows --0.19 dB---both exhibiting similar degradation patterns despite different matching strategies. This indicates that the architectural benefit of separation is fundamental to the optimization dynamics, not an artifact of any particular pipeline configuration. This validates our architectural choice of maintaining separate parameters for two reasons:

\begin{enumerate}
    \item \textbf{Conflicting gradient signals.} Gaussian means receive gradients from photometric rendering that optimize for appearance quality---pushing primitives to minimize $\ell_1$ and SSIM losses. Track points receive gradients from reprojection that enforce multi-view geometric consistency. When merged, these objectives compete: a Gaussian may need to move for better rendering but should stay fixed for geometric anchoring, creating optimization conflicts that degrade both.
    
    \item \textbf{Densification disrupts geometric anchors.} MCMC densification performs stochastic birth-death processes that relocate, split, and merge Gaussians based on rendering quality. If track points were tied to Gaussian means, densification would inadvertently modify the geometric anchors, breaking the multi-view consistency constraints. Separate track points remain stable throughout training regardless of how Gaussians evolve.
\end{enumerate}

This ablation confirms that the separation is not merely an implementation detail but a principled architectural choice that enables photometric and geometric objectives to coexist without interference.

\section{Algorithm: Joint Optimization Loop}
\label{sec:algorithm_appendix}

\Cref{alg:joint_opt} presents the complete joint optimization loop that distinguishes GloSplat from prior work. The key architectural novelty is maintaining \textbf{track 3D points} $\{\mathbf{X}_k\}$ as \emph{separate optimizable parameters} from Gaussian means $\{\boldsymbol{\mu}_j\}$. This enables the reprojection-based BA loss to provide geometric anchoring independent of Gaussian rendering quality, preventing early-stage pose drift while photometric gradients enable fine-grained refinement.

\begin{algorithm}[h]
\caption{GloSplat Joint Optimization Loop}
\label{alg:joint_opt}
\begin{algorithmic}[1]
\REQUIRE Images $\{\mathbf{I}_i\}_{i=1}^{N}$, SfM points $\{\mathbf{P}_k\}$, feature tracks $\{\mathcal{T}_k\}$, initial poses $\{\mathbf{T}_i^{(0)}\}$, intrinsics $\{\mathbf{K}_i\}$
\ENSURE Optimized Gaussians $\mathcal{G}$, refined poses $\{\mathbf{T}_i\}$, refined track points $\{\mathbf{X}_k\}$

\STATE \textbf{// Initialize from Global SfM}
\STATE Initialize Gaussians $\mathcal{G} \leftarrow \{\boldsymbol{\mu}_j, \boldsymbol{\Sigma}_j, \alpha_j, \mathbf{c}_j\}$ from SfM points
\STATE Initialize track points $\{\mathbf{X}_k\} \leftarrow \{\mathbf{P}_k\}$ \COMMENT{Separate from $\boldsymbol{\mu}_j$}
\STATE Initialize pose adjustments $\{\Delta\mathbf{T}_i\} \leftarrow \mathbf{0}$
\STATE

\FOR{$t = 1$ to $T_{\max}$}
    \STATE \textbf{// Sample training batch}
    \STATE Sample image indices $\mathcal{B} \subset \{1, \ldots, N\}$
    \STATE

    \FOR{$i \in \mathcal{B}$}
        \STATE \textbf{// Apply pose adjustment}
        \STATE $\mathbf{T}_i \leftarrow \mathbf{T}_i^{(0)} \oplus \Delta\mathbf{T}_i$ \COMMENT{SE(3) composition}
        \STATE

        \STATE \textbf{// Render via 3D Gaussian Splatting}
        \STATE $\hat{\mathbf{I}}_i \leftarrow \text{Rasterize}(\mathcal{G}, \mathbf{T}_i, \mathbf{K}_i)$
        \STATE

        \STATE \textbf{// Photometric loss}
        \STATE $\mathcal{L}_{\text{photo}} \leftarrow (1-\lambda_{\text{SSIM}})\|\hat{\mathbf{I}}_i - \mathbf{I}_i\|_1 + \lambda_{\text{SSIM}}(1 - \text{SSIM}(\hat{\mathbf{I}}_i, \mathbf{I}_i))$
    \ENDFOR
    \STATE

    \STATE \textbf{// Joint BA loss on track points (key distinction)}
    \STATE $\mathcal{L}_{\text{BA}} \leftarrow 0$
    \FOR{each track $\mathcal{T}_k$ with observations in $\mathcal{B}$}
        \FOR{each observation $(i, \mathbf{x}_{i,p}) \in \mathcal{T}_k$}
            \STATE $\hat{\mathbf{x}}_{i,p} \leftarrow \pi(\mathbf{K}_i, \mathbf{T}_i, \mathbf{X}_k)$ \COMMENT{Project track point}
            \STATE $\mathcal{L}_{\text{BA}} \leftarrow \mathcal{L}_{\text{BA}} + \rho_{\text{Huber}}(\|\hat{\mathbf{x}}_{i,p} - \mathbf{x}_{i,p}\|^2; \delta)$
        \ENDFOR
    \ENDFOR
    \STATE

    \STATE \textbf{// Combined loss}
    \STATE $\mathcal{L} \leftarrow \mathcal{L}_{\text{photo}} + \lambda_{\text{BA}} \mathcal{L}_{\text{BA}}$
    \STATE

    \STATE \textbf{// Backward and update all parameters jointly}
    \STATE Backpropagate $\nabla\mathcal{L}$
    \STATE Update Gaussians: $\mathcal{G} \leftarrow \mathcal{G} - \eta_{\mathcal{G}} \nabla_{\mathcal{G}}\mathcal{L}$
    \STATE Update poses: $\Delta\mathbf{T}_i \leftarrow \Delta\mathbf{T}_i - \eta_{\text{pose}} \nabla_{\Delta\mathbf{T}_i}\mathcal{L}$ \COMMENT{Receives both gradients}
    \STATE Update track points: $\mathbf{X}_k \leftarrow \mathbf{X}_k - \eta_{\text{BA}} \nabla_{\mathbf{X}_k}\mathcal{L}_{\text{BA}}$
    \STATE

    \STATE \textbf{// MCMC densification (adopted from prior work)}
    \STATE $\mathcal{G} \leftarrow \text{MCMCDensify}(\mathcal{G}, \text{gradients})$
\ENDFOR

\STATE \textbf{return} $\mathcal{G}$, $\{\mathbf{T}_i^{(0)} \oplus \Delta\mathbf{T}_i\}$, $\{\mathbf{X}_k\}$
\end{algorithmic}
\end{algorithm}

\paragraph{Key Implementation Details.}
\begin{itemize}
    \item \textbf{Separate track points (Lines 3, 22--27):} Unlike prior joint methods that only optimize Gaussian means $\boldsymbol{\mu}_j$, we maintain explicit track 3D points $\mathbf{X}_k$ as separate parameters. This enables geometric constraints via reprojection loss independent of Gaussian rendering quality.
    \item \textbf{Dual gradient flow (Lines 30--32):} Camera poses receive gradients from \emph{both} photometric loss (via differentiable rendering) and BA loss (via reprojection). The BA loss provides geometric anchoring that stabilizes early training when Gaussians are sparse.
    \item \textbf{Projection function (Line 24):} $\pi(\mathbf{K}, \mathbf{T}, \mathbf{X}) = \text{proj}(\mathbf{K}(\mathbf{R}\mathbf{X} + \mathbf{t}))$ where $\mathbf{T} = [\mathbf{R}|\mathbf{t}]$ and $\text{proj}([x,y,z]^\top) = [x/z, y/z]^\top$.
    \item \textbf{Hyperparameters:} $\lambda_{\text{SSIM}} = 0.2$, $\lambda_{\text{BA}} = 10^{-4}$, $\delta = 1.0$ (Huber threshold), $\eta_{\text{pose}} = 10^{-5}$, $T_{\max} = 30000$.
\end{itemize}
The full pipeline consists of three stages: (1) \texttt{glo-feat} for feature extraction and matching (frozen preprocessing), (2) \texttt{glo-sfm} for global SfM initialization (rotation averaging, positioning, bundle adjustment), and (3) \texttt{glo-joint} for joint 3DGS training with the algorithm above.

\end{document}